\documentclass[10pt,twocolumn,letterpaper]{article}
\usepackage[pagenumbers]{cvpr}
\usepackage[T1]{fontenc}
\usepackage{microtype,array,longtable,tikz,placeins}
\usetikzlibrary{arrows.meta,positioning,fit}
\definecolor{cvprblue}{rgb}{0.21,0.49,0.74}
\usepackage[pagebackref,breaklinks,colorlinks,allcolors=cvprblue]{hyperref}
\def\confName{CVPR}
\def\confYear{2027}
\newcounter{algorithm}
\newcommand{\vpr}{\textsc{VPR}}

\newcommand{\R}{\mathbb{R}}
\newcommand{\imgset}{\mathcal{I}}
\newcommand{\rot}{R_{\Theta}}

\DeclareMathOperator*{\argmax}{arg\,max}

\title{Keep Images in Sight: Visual Position Re-Anchoring\\for Long-Context MLLMs}
\author{Juntao Xu\textsuperscript{*} \quad Qichai Zhao\textsuperscript{*} \quad Rui Sun \quad Zhouyang Jiang\\[3pt]
Shiwei Wu \quad Zhi Li \quad Kaisheng Ma \quad Yu Liu\textsuperscript{\textdagger}\\[4pt]
Tsinghua University\\[3pt]
{\small \textsuperscript{*}Equal contribution. \quad \textsuperscript{\textdagger}Corresponding author.}}
\date{}
\hypersetup{
  pdftitle={Keep Images in Sight: Visual Position Re-Anchoring for Long-Context MLLMs},
  pdfauthor={Juntao Xu, Qichai Zhao, Rui Sun, Zhouyang Jiang, Shiwei Wu, Zhi Li, Kaisheng Ma, Yu Liu}
}
\begin{document}
\raggedbottom
\setlength{\textfloatsep}{10pt plus 2pt minus 2pt}
\setlength{\dbltextfloatsep}{10pt plus 2pt minus 2pt}
\setlength{\floatsep}{8pt plus 2pt minus 2pt}
\setlength{\intextsep}{10pt plus 2pt minus 2pt}
\widowpenalty=10000
\clubpenalty=10000
\captionsetup[table]{skip=3pt}
\maketitle
\begin{abstract}
Long-context multimodal intelligence requires revisiting earlier images as conversations and documents accumulate, yet retaining an image does not guarantee reliable access to its details. By varying context length, redistributing a fixed text passage around the image, and changing positions without adding text, we identify the positional relation between images and questions as a direct influence on visual access. To address this gap, we propose Visual Position Re-Anchoring (VPR), a training-free method that shifts cached visual-key rotary phases toward a short-context reference before question processing. It operates directly on the existing KV cache, preserving standard attention computation without extracting attention maps or adding model forward passes. We develop static and continuous compensation schedules, using the static schedule by default. Across ten checkpoints and ten benchmarks, static VPR improves 97 of 100 model--task scores. Further analysis shows greater attention to regions containing answer evidence across six checkpoints, with minimal runtime overhead under fixed rotary frequencies. These findings demonstrate an efficient way to sustain visual access through positional adjustment within the cache.

\end{abstract}
\vspace{-6pt}
\section{Introduction}
\label{sec:intro}
\begin{figure}[!t]
\centering
\captionsetup{skip=3pt}
\includegraphics[trim=0bp 2bp 20bp 0bp,clip,width=.92\columnwidth]{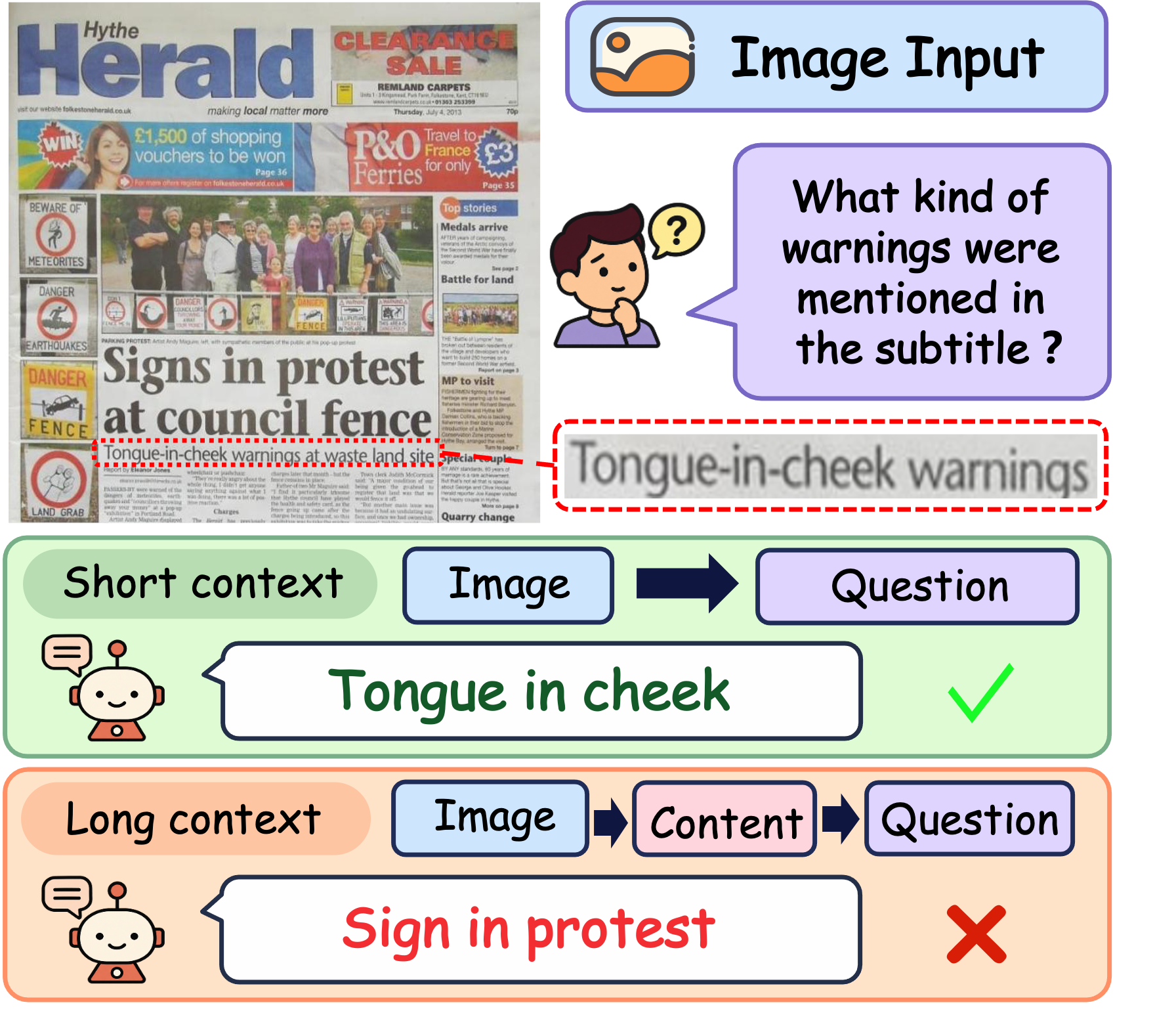}
\caption{\textbf{Visual evidence drift in long contexts.} The correct short-context answer, ``Tongue in cheek,'' becomes ``Sign in protest'' in a long context, confusing the subtitle with the headline.}
\label{fig:intro_case}
\end{figure}
\begin{figure*}[t]
\centering
\captionsetup{skip=3pt}
\includegraphics[width=.99\textwidth,trim={8.9455bp 10.9091bp 0bp 1.3091bp},clip]{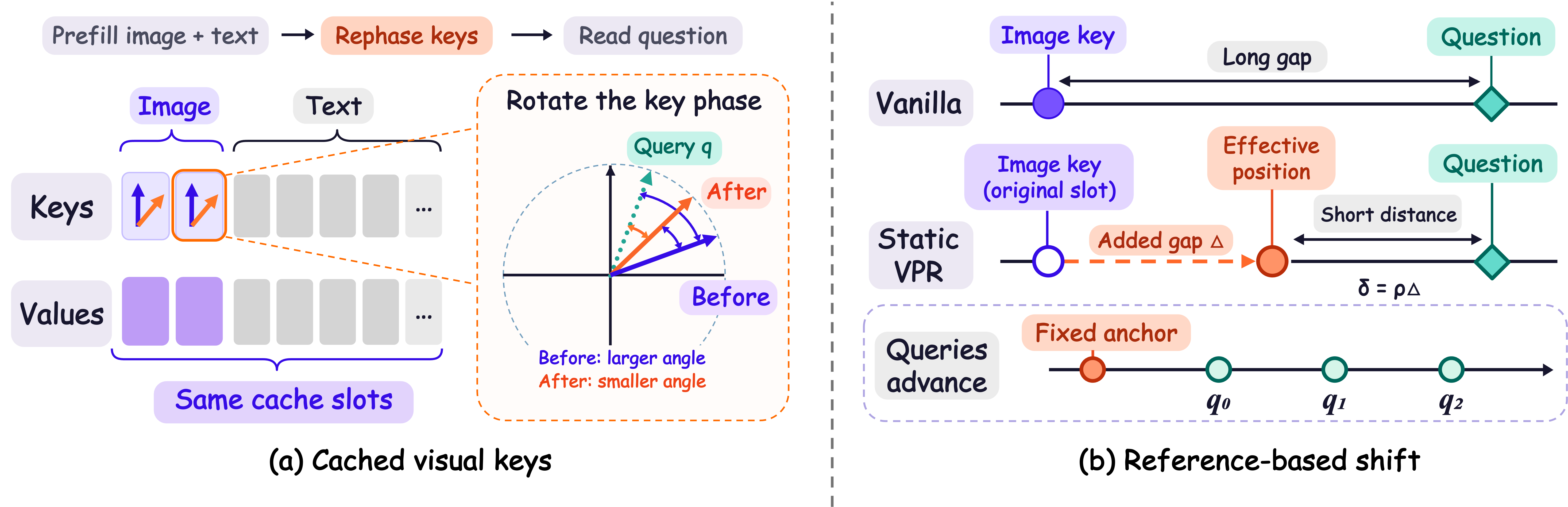}
\caption{Overview of VPR. (a) Left: image keys are rephased before question processing; values and cache slots are preserved. One rotary pair is shown schematically. (b) Right: $\rho=1$ restores the initial short context separation. The offset stays fixed as decoding queries advance.}
\label{fig:overview}
\end{figure*}

Multimodal large language models (MLLMs) enable applications such as personal assistance, document management, and knowledge integration, where users work with images alongside text. As their capabilities expand, these applications call for increasingly complex interactions that extend beyond a single visual question and answer. A user may share an image early in a conversation, review several documents or exchange further messages, and only then return to a detail in the original image. Supporting such interactions requires sustained access to visual evidence throughout an evolving context. Recent models, including Qwen and InternVL, support long multimodal inputs~\citep{bai2025qwen25vl,bai2025qwen3vl,zhu2025internvl3,wang2025internvl35}, but retaining an image within the context window does not by itself ensure that the model can still use it reliably. How well can MLLMs access earlier visual evidence as the intervening text grows?

Figure~\ref{fig:intro_case} illustrates this challenge on a newspaper page. Asked about the warnings in the subtitle, the model answers correctly in a short context but echoes the headline after a long passage. We refer to this failure as \emph{visual evidence drift}: the answer shifts away from the visual detail required by the question, even though the image remains available in the context. To examine how visual access changes with context length, we hold the image and question fixed while varying the amount of text between them. For example, Qwen2.5-VL-7B's DocVQA~\citep{mathew2020docvqa} score falls from 93.5 in the short context to 52.6 ANLS points with 32K intervening text tokens, even though the input fits within the checkpoint's configured context window. This gap between retaining an image and using its evidence motivates our investigation.

Existing work strengthens visual grounding through attention intervention~\citep{liu2024pai,zhang2024eah,zhao2026vga}, decoding-time correction~\citep{leng2023vcd,huang2024opera,chen2024halc}, visual revisitation~\citep{sun2025tvc}, and positional design~\citep{ge2025v2pe,chen2026dipe}. Depending on the design, these approaches can require training adaptation, intermediate-attention processing, auxiliary predictions, or repeated visual-token processing. For interventions applied throughout decoding, the additional operations accumulate over generation steps, motivating an efficient way to restore visual access while retaining standard attention computation.

We investigate this failure through three complementary diagnostics. First, length sweeps reveal reduced answer accuracy and image-attention allocation in several models, with degradation patterns that depend on the checkpoint and context length. Second, redistributing the same text before and after the image across five layouts changes accuracy at a fixed total token budget, showing that context size alone does not explain the failure. Third, shifting question positions forward or cached visual-key phases backward on otherwise identical short inputs can lower task scores without adding competing text keys, establishing positional relations as a direct influence on visual access. Together, these findings motivate re-anchoring visual-key phases while retaining the full text context.

These findings suggest a direct way to improve access to visual evidence already stored during inference. Long-context generation uses a key--value (KV) cache to retain and reuse prefix representations~\citep{kwon2023pagedattention}; the cached visual keys therefore provide a point of intervention without recomputing the image or the prefix. We introduce Visual Position Re-Anchoring (\vpr), a training-free method that directly adjusts these keys' rotary phases to improve their interaction with subsequent queries. After context prefill and before question processing, \vpr\ shifts the visual-key phases toward a short-context reference, preserving visual values, text-cache entries, and physical cache slots. Our static schedule holds this offset fixed during decoding, while our continuous schedule updates it as each new query advances.

VPR improves answer performance across ten checkpoints and ten benchmarks, while evidence-region analysis shows greater attention to answer-bearing regions in six checkpoints. Static VPR achieves these gains with little additional runtime cost and the lowest overhead among the compared attention interventions. These results establish cache-level positional adjustment as a practical approach to sustaining access to earlier visual evidence.

Our contributions are threefold:
\begin{itemize}
\item \textbf{Diagnosing long-context visual access.} We combine length sweeps, fixed-budget image layouts, and position-only controls to establish the role of positional relations in accessing earlier visual evidence.
\item \textbf{Efficient visual re-anchoring.} We introduce a training-free, plug-and-play intervention on cached visual keys. Its static and continuous schedules reuse the computed prefix and retain standard attention computation, without attention-map extraction or additional model forward passes.
\item \textbf{Broad empirical validation.} Experiments across ten checkpoints and ten benchmarks demonstrate improved long-context visual grounding, supported by evidence-region attention analysis and runtime measurements.
\end{itemize}

\section{Related Work}
\label{sec:related}
\noindent\textbf{Visual evidence in extended contexts.}\enspace
Visual instruction tuning connects images to language-model generation~\citep{liu2023llava}, but extended reasoning can compromise perceptual fidelity~\citep{liu2025morethinking}. VisualSwap tests whether models revisit changed visual evidence~\citep{shi2026visualswap}. TVC combines visual-conditioning training with compressed visual-token reinjection and cache resets during reasoning~\citep{sun2025tvc}. We study a complementary setting: a fixed image queried after a long passage, with intervention on its existing cache.

\par\addvspace{4pt}
\noindent\textbf{Attention-based visual enhancement.}\enspace
PAI increases visual attention logits and adds a text-only model branch for contrastive decoding~\citep{liu2024pai}. EAH selects a shallow-layer head by its visual attention pattern and broadcasts its attention matrix to other heads; its authors report longer decoding than greedy inference~\citep{zhang2024eah}. These interventions access intermediate attention tensors, with PAI additionally evaluating a second branch. VPR modifies cached-key phases, avoiding attention-map extraction, head selection and auxiliary decoding. Section~\ref{sec:efficiency} measures its cache-update cost.

\par\addvspace{4pt}
\noindent\textbf{Rotary positions and cross-modal anchoring.}\enspace
RoPE encodes relative positions through query--key rotations~\citep{su2021roformer}. Analyses challenge a simple distance-decay account~\citep{barbero2025round}, motivating our controlled positional interventions. CCA trains with concentric visual positions and causal masks to shorten visual--instruction distances~\citep{xing2024cca}. MRoPE variants differ in their axis-to-dimension assignments~\citep{bai2025qwen25vl,bai2025qwen3vl}. DIPE anchors cross-modal queries to their modality segment's first position, retaining native same-modality positions~\citep{chen2026dipe}. VPR instead rephases cached visual keys after prefix computation in frozen models, preserving native visual-coordinate structure and causal masks. It directly targets stored evidence without changing query positions. Section~\ref{sec:query_anchoring} derives the relative-phase connection to query anchoring.

\section{Diagnosing Visual Access in Long Contexts}
\label{sec:diagnosis}
\subsection{Visual access across context lengths}
\label{sec:length_diagnosis}
We evaluate $[I,C_L,Q]$, where $I$ is an image, $Q$ requires visual evidence, and $C_L$ contains $L\in\{0,1,2,4,8,16,32\}$K intervening text tokens ($K=1{,}024$). Vanilla and VPR receive identical inputs through native chat templates. This input length is distinct from the answer budget; Section~\ref{sec:setup} gives the protocol.

\begin{figure}[t]\centering
\includegraphics[width=\columnwidth]{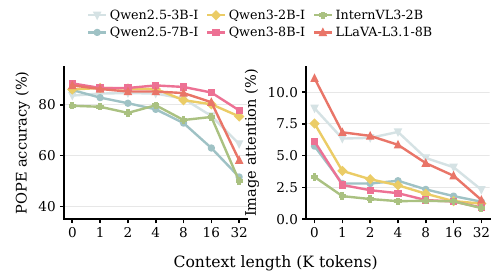}
\caption{Visual access across context lengths. Vanilla POPE accuracy and image-attention mass measure task performance and allocation, respectively; they use separate cohorts (Appendix~\ref{app:pope_attention}).}
\label{fig:length}\end{figure}

We measure visual access through answer accuracy and attention allocation. For the final prompt query predicting the first answer token, the head-averaged image-attention mass in layer $\ell$ is
\begin{equation}
A_\ell=\frac{1}{H_\ell}\sum_{h=1}^{H_\ell}\sum_{i\in\imgset}
\operatorname{softmax}(s^{\ell,h}_{\mathrm{readout}})_i,
\label{eq:attention}
\end{equation}
where $\imgset$ indexes visual tokens, $H_\ell$ counts query heads, and softmax normalizes over all available keys. Figure~\ref{fig:length} averages over layers and examples from a fixed attention cohort across lengths, separate from the accuracy cohort. This mass measures allocation, not evidence correctness; Section~\ref{sec:attention_results} examines annotated evidence regions.

Figure~\ref{fig:length} shows substantial accuracy losses in several models alongside reduced image attention. Between 0K and 32K, LLaVA's image-attention mass falls from 11.08\% to 1.51\%, and Qwen2.5-7B's from 5.74\% to 1.38\%. Accuracy varies non-monotonically across checkpoints: Qwen3-8B-Instruct scores 84.8\% at 16K and 77.8\% at 32K. These patterns do not imply a universal distance-decay law or make attention mass a direct measure of understanding.

\subsection{Equal-length layouts reveal position sensitivity}
\label{sec:layout_diagnosis}
\begin{figure}[!t]\centering
\captionsetup{skip=3pt}
{\small\normalfont (a) Input layouts\par}
\vspace{1pt}
\includegraphics[width=\columnwidth]{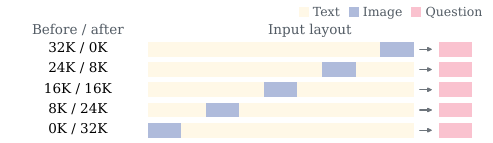}\par
\vspace{3pt}
{\small\normalfont (b) POPE accuracy\par}
\vspace{1pt}
\includegraphics[trim=0bp 0bp 0bp 12bp,clip,width=\columnwidth]{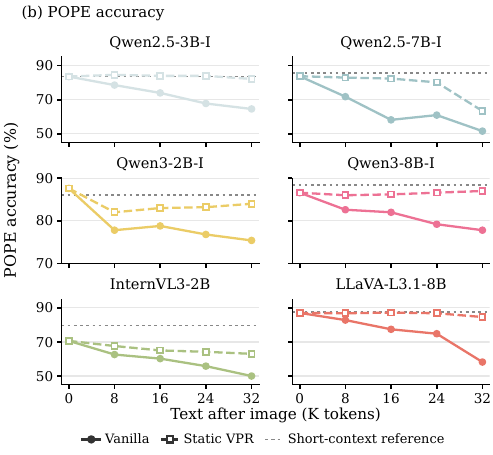}
\caption{Image position at a fixed 32K text budget. (a) The same text is split before and after the image. (b) Dotted lines mark each model's 0K POPE accuracy from Figure~\ref{fig:length}. Middle-row panels use 70--90\% accuracy; other panels use 45--95\%.}
\label{fig:layout_scan}\end{figure}

The length sweep reveals weaker visual access but also changes competing keys and prefix representations. We therefore test whether image placement affects accuracy within the same long context. We split a fixed 32K-token passage into $[C_{\mathrm{before}},I,C_{\mathrm{after}},Q]$, allocating 0, 8, 16, 24, or 32K tokens after the image and the remainder before it. The image and question remain fixed across all five layouts. These layouts preserve text content, token order, and total budget. Within each layout, Vanilla and VPR receive identical inputs, as shown in Figure~\ref{fig:layout_scan}.

Several checkpoints perform better with more text before the image despite the unchanged budget. Visual access thus depends on evidence placement beyond text quantity alone. Section~\ref{sec:layout_recovery} analyzes VPR under these layouts.

\subsection{Position changes without added text}
\label{sec:position_control}
\begin{figure}[!t]\centering
\includegraphics[width=\columnwidth]{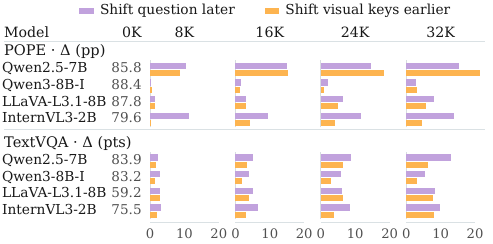}
\caption{Position shifts on identical short inputs. Bars show the score loss relative to each model's unshifted score (0K) as the question moves later or the visual keys move earlier.}
\label{fig:position_control}\end{figure}

Image placement affects accuracy at a fixed text budget, but also changes the prefix shaping image representations. We next test whether this sensitivity can also arise directly from positional encoding, independently of the intervening passage. We hold the short image--question input and causal visibility fixed. Figure~\ref{fig:position_control} compares shifting the question-processing segment forward with rephasing cached visual keys backward before question processing. Offsets reproduce the added separation of 8--32K settings without processing intervening text. The former changes question positions relative to the entire prefix; the latter targets visual keys. The plotted $\Delta$ subtracts shifted from unshifted scores, in percentage points (pp) for POPE and consensus-score points (pts) for TextVQA.

On POPE, an 8K question shift lowers InternVL3-2B accuracy by 11.2 percentage points, versus 0.4 for the key shift. On TextVQA, both 24K shifts lower all four models' scores: InternVL3-2B loses 8.5 and 3.7 points, respectively, whereas LLaVA loses about six under either. Rotary positional changes can therefore impair visual access without additional text keys. Together, these three experiments identify positional relations as a direct intervention target, motivating re-anchoring with the full long context retained.

\section{Visual Position Re-Anchoring}
\label{sec:method}
Figure~\ref{fig:overview} illustrates \vpr: rephasing cached visual keys directly adjusts their positional relations to later queries, improving access to earlier visual evidence.

\subsection{Rephasing cached visual keys}
For one head, let $\bar q_m,\bar k_n\in\R^{d_h}$ denote unrotated content vectors. With fixed rotary frequencies $\Theta$, $q_m=\rot(m)\bar q_m$ and $k_n=\rot(n)\bar k_n$. Their attention logit is
\begin{equation}
s_{m,n}=\frac{\bar q_m^\top\rot(n-m)\bar k_n}{\sqrt{d_h}}.
\label{eq:rope}
\end{equation}
Relative phase modulates query--key compatibility without requiring monotonic distance decay. The RoPE composition law enables a direct cache intervention:
\begin{equation}
\widetilde k_n=\rot(\delta)k_n=\rot(n+\delta)\bar k_n.
\label{eq:shift}
\end{equation}
A positive $\delta$ moves the key's effective position toward a later question. We apply this transformation to image keys at every decoder layer and KV head. Visual values, text-cache entries, physical slots and causal masks are unchanged. Subsequent question representations adapt through attention to the modified keys.

\subsection{Choosing a short-context reference}
We obtain the reference positions by placing the same image and question in the native chat template with no intervening passage. Let $p_Q$ be the first question token's native position, $a$ the first native position after the image, and $b_0$ their separation when the same image and question are adjacent under the native template. Define the extra input gap and compensation as
\begin{equation}
\Delta=(p_Q-a)-b_0,\qquad\delta=\rho\Delta.
\label{eq:delta}
\end{equation}
Here $\rho\geq0$ is a dimensionless compensation strength: $\rho=0$ leaves the cache unchanged, and $\rho=1$ removes the full added gap. The reference requires only prompt preparation, not an additional model forward pass. The offset uses prepared position IDs rather than the nominal text length. It can therefore accommodate image expansion and template tokens without assuming one rotary position per input-array entry.

For image token $i$, define signed separation as $d_i=p_Q-p_i$, with $d_i^{(0)}$ its short-context counterpart. With the image's native coordinate structure unchanged, the long input adds $\Delta$; re-anchoring yields
\begin{equation}
\widetilde d_i=d_i^{(0)}+(1-\rho)\Delta.
\label{eq:distance}
\end{equation}
At the default $\rho=1$, the added separation is compensated. Strengths above one shift beyond the reference and reverse the signed separation when $\rho>1+d_i^{(0)}/\Delta$ for $\Delta>0$. This reversal concerns effective positions, not causal visibility. For multidimensional RoPE, this relation holds per coordinate axis. Applying the same offset to every visual key preserves the relative coordinates of image tokens while changing their rotary phases relative to the question.

\begin{table*}[t]\centering\fontsize{7.7}{8.4}\selectfont
\setlength{\tabcolsep}{1pt}\renewcommand{\arraystretch}{1.03}\setlength{\aboverulesep}{1.4pt}\setlength{\belowrulesep}{1.4pt}
\caption{Ten checkpoints across ten visual tasks at 32K. Parentheses show VPR gains: red for increases, green for decreases. DocVQA uses ANLS, TextVQA uses consensus scores, and other tasks use accuracy, all on a 0--100 scale. I/T: Instruct/Thinking; L3.1/OV: LLaVA-Llama3.1/OneVision. $\dagger$Context extrapolation.}\label{tab:main}
\begin{tabular*}{\textwidth}{@{\extracolsep{\fill}}llcccccccccc@{}}\toprule
Task & Method & \multicolumn{2}{c}{\textbf{Qwen2.5-VL}} & \multicolumn{4}{c}{\textbf{Qwen3-VL}} & \multicolumn{2}{c}{\textbf{InternVL3}} & \multicolumn{2}{c}{\textbf{LLaVA}} \\
 & & \multicolumn{2}{c}{MRoPE} & \multicolumn{4}{c}{MRoPE-I} & \multicolumn{2}{c}{1D + NTK} & \multicolumn{2}{c}{1D: L3.1 / Qwen2} \\
\cmidrule(lr){3-4}\cmidrule(lr){5-8}\cmidrule(lr){9-10}\cmidrule(lr){11-12}
 & & 3B-I & 7B-I & 2B-I & 4B-I & 8B-I & 8B-T & 2B & 8B & L3.1-8B & OV-7B\textdagger \\\midrule
POPE & Vanilla & 64.6 & 51.6 & 75.4 & 85.0 & 77.8 & 82.0 & 50.0 & 79.0 & 58.2 & 64.4 \\
 & \vpr & 82.2\,{\fontsize{6.5}{7.0}\selectfont\textcolor[rgb]{0.71,0.12,0.18}{(+17.6)}} & 63.4\,{\fontsize{6.5}{7.0}\selectfont\textcolor[rgb]{0.71,0.12,0.18}{(+11.8)}} & 84.0\,{\fontsize{6.5}{7.0}\selectfont\textcolor[rgb]{0.71,0.12,0.18}{(+8.6)}} & 86.0\,{\fontsize{6.5}{7.0}\selectfont\textcolor[rgb]{0.71,0.12,0.18}{(+1.0)}} & 87.0\,{\fontsize{6.5}{7.0}\selectfont\textcolor[rgb]{0.71,0.12,0.18}{(+9.2)}} & 83.0\,{\fontsize{6.5}{7.0}\selectfont\textcolor[rgb]{0.71,0.12,0.18}{(+1.0)}} & 63.0\,{\fontsize{6.5}{7.0}\selectfont\textcolor[rgb]{0.71,0.12,0.18}{(+13.0)}} & 84.0\,{\fontsize{6.5}{7.0}\selectfont\textcolor[rgb]{0.71,0.12,0.18}{(+5.0)}} & 84.6\,{\fontsize{6.5}{7.0}\selectfont\textcolor[rgb]{0.71,0.12,0.18}{(+26.4)}} & 83.8\,{\fontsize{6.5}{7.0}\selectfont\textcolor[rgb]{0.71,0.12,0.18}{(+19.4)}} \\
\midrule
CountBench & Vanilla & 43.6 & 44.2 & 64.0 & 64.0 & 59.0 & 79.0 & 5.0 & 57.0 & 36.6 & 62.0 \\
 & \vpr & 72.0\,{\fontsize{6.5}{7.0}\selectfont\textcolor[rgb]{0.71,0.12,0.18}{(+28.4)}} & 71.2\,{\fontsize{6.5}{7.0}\selectfont\textcolor[rgb]{0.71,0.12,0.18}{(+27.0)}} & 68.0\,{\fontsize{6.5}{7.0}\selectfont\textcolor[rgb]{0.71,0.12,0.18}{(+4.0)}} & 69.0\,{\fontsize{6.5}{7.0}\selectfont\textcolor[rgb]{0.71,0.12,0.18}{(+5.0)}} & 67.0\,{\fontsize{6.5}{7.0}\selectfont\textcolor[rgb]{0.71,0.12,0.18}{(+8.0)}} & 86.0\,{\fontsize{6.5}{7.0}\selectfont\textcolor[rgb]{0.71,0.12,0.18}{(+7.0)}} & 32.0\,{\fontsize{6.5}{7.0}\selectfont\textcolor[rgb]{0.71,0.12,0.18}{(+27.0)}} & 58.0\,{\fontsize{6.5}{7.0}\selectfont\textcolor[rgb]{0.71,0.12,0.18}{(+1.0)}} & 54.4\,{\fontsize{6.5}{7.0}\selectfont\textcolor[rgb]{0.71,0.12,0.18}{(+17.8)}} & 76.8\,{\fontsize{6.5}{7.0}\selectfont\textcolor[rgb]{0.71,0.12,0.18}{(+14.8)}} \\
\midrule
ChartQA & Vanilla & 37.4 & 17.2 & 32.0 & 45.0 & 54.0 & 39.0 & 1.0 & 44.0 & 12.2 & 50.0 \\
 & \vpr & 74.8\,{\fontsize{6.5}{7.0}\selectfont\textcolor[rgb]{0.71,0.12,0.18}{(+37.4)}} & 33.4\,{\fontsize{6.5}{7.0}\selectfont\textcolor[rgb]{0.71,0.12,0.18}{(+16.2)}} & 51.0\,{\fontsize{6.5}{7.0}\selectfont\textcolor[rgb]{0.71,0.12,0.18}{(+19.0)}} & 59.0\,{\fontsize{6.5}{7.0}\selectfont\textcolor[rgb]{0.71,0.12,0.18}{(+14.0)}} & 72.0\,{\fontsize{6.5}{7.0}\selectfont\textcolor[rgb]{0.71,0.12,0.18}{(+18.0)}} & 56.0\,{\fontsize{6.5}{7.0}\selectfont\textcolor[rgb]{0.71,0.12,0.18}{(+17.0)}} & 7.0\,{\fontsize{6.5}{7.0}\selectfont\textcolor[rgb]{0.71,0.12,0.18}{(+6.0)}} & 54.0\,{\fontsize{6.5}{7.0}\selectfont\textcolor[rgb]{0.71,0.12,0.18}{(+10.0)}} & 17.4\,{\fontsize{6.5}{7.0}\selectfont\textcolor[rgb]{0.71,0.12,0.18}{(+5.2)}} & 67.8\,{\fontsize{6.5}{7.0}\selectfont\textcolor[rgb]{0.71,0.12,0.18}{(+17.8)}} \\
\midrule
MathVista & Vanilla & 29.0 & 38.4 & 40.0 & 51.0 & 61.0 & 52.0 & 38.8 & 38.0 & 23.8 & 47.0 \\
 & \vpr & 42.4\,{\fontsize{6.5}{7.0}\selectfont\textcolor[rgb]{0.71,0.12,0.18}{(+13.4)}} & 47.4\,{\fontsize{6.5}{7.0}\selectfont\textcolor[rgb]{0.71,0.12,0.18}{(+9.0)}} & 44.0\,{\fontsize{6.5}{7.0}\selectfont\textcolor[rgb]{0.71,0.12,0.18}{(+4.0)}} & 60.0\,{\fontsize{6.5}{7.0}\selectfont\textcolor[rgb]{0.71,0.12,0.18}{(+9.0)}} & 63.0\,{\fontsize{6.5}{7.0}\selectfont\textcolor[rgb]{0.71,0.12,0.18}{(+2.0)}} & 59.0\,{\fontsize{6.5}{7.0}\selectfont\textcolor[rgb]{0.71,0.12,0.18}{(+7.0)}} & 44.6\,{\fontsize{6.5}{7.0}\selectfont\textcolor[rgb]{0.71,0.12,0.18}{(+5.8)}} & 43.0\,{\fontsize{6.5}{7.0}\selectfont\textcolor[rgb]{0.71,0.12,0.18}{(+5.0)}} & 26.8\,{\fontsize{6.5}{7.0}\selectfont\textcolor[rgb]{0.71,0.12,0.18}{(+3.0)}} & 51.0\,{\fontsize{6.5}{7.0}\selectfont\textcolor[rgb]{0.71,0.12,0.18}{(+4.0)}} \\
\midrule
DocVQA & Vanilla & 57.0 & 52.6 & 53.6 & 72.0 & 80.4 & 83.6 & 2.8 & 26.0 & 20.6 & 66.8 \\
 & \vpr & 81.8\,{\fontsize{6.5}{7.0}\selectfont\textcolor[rgb]{0.71,0.12,0.18}{(+24.8)}} & 76.8\,{\fontsize{6.5}{7.0}\selectfont\textcolor[rgb]{0.71,0.12,0.18}{(+24.2)}} & 66.8\,{\fontsize{6.5}{7.0}\selectfont\textcolor[rgb]{0.71,0.12,0.18}{(+13.2)}} & 72.8\,{\fontsize{6.5}{7.0}\selectfont\textcolor[rgb]{0.71,0.12,0.18}{(+0.8)}} & 83.0\,{\fontsize{6.5}{7.0}\selectfont\textcolor[rgb]{0.71,0.12,0.18}{(+2.6)}} & 85.8\,{\fontsize{6.5}{7.0}\selectfont\textcolor[rgb]{0.71,0.12,0.18}{(+2.2)}} & 35.6\,{\fontsize{6.5}{7.0}\selectfont\textcolor[rgb]{0.71,0.12,0.18}{(+32.8)}} & 27.6\,{\fontsize{6.5}{7.0}\selectfont\textcolor[rgb]{0.71,0.12,0.18}{(+1.6)}} & 32.0\,{\fontsize{6.5}{7.0}\selectfont\textcolor[rgb]{0.71,0.12,0.18}{(+11.4)}} & 76.8\,{\fontsize{6.5}{7.0}\selectfont\textcolor[rgb]{0.71,0.12,0.18}{(+10.0)}} \\
\midrule
GQA & Vanilla & 38.6 & 18.4 & 3.0 & 35.0 & 45.0 & 25.0 & 0.0 & 6.0 & 42.6 & 40.2 \\
 & \vpr & 54.8\,{\fontsize{6.5}{7.0}\selectfont\textcolor[rgb]{0.71,0.12,0.18}{(+16.2)}} & 16.6\,{\fontsize{6.5}{7.0}\selectfont\textcolor[rgb]{0.13,0.52,0.28}{(-1.8)}} & 17.0\,{\fontsize{6.5}{7.0}\selectfont\textcolor[rgb]{0.71,0.12,0.18}{(+14.0)}} & 40.0\,{\fontsize{6.5}{7.0}\selectfont\textcolor[rgb]{0.71,0.12,0.18}{(+5.0)}} & 47.0\,{\fontsize{6.5}{7.0}\selectfont\textcolor[rgb]{0.71,0.12,0.18}{(+2.0)}} & 46.0\,{\fontsize{6.5}{7.0}\selectfont\textcolor[rgb]{0.71,0.12,0.18}{(+21.0)}} & 4.0\,{\fontsize{6.5}{7.0}\selectfont\textcolor[rgb]{0.71,0.12,0.18}{(+4.0)}} & 7.0\,{\fontsize{6.5}{7.0}\selectfont\textcolor[rgb]{0.71,0.12,0.18}{(+1.0)}} & 61.8\,{\fontsize{6.5}{7.0}\selectfont\textcolor[rgb]{0.71,0.12,0.18}{(+19.2)}} & 55.4\,{\fontsize{6.5}{7.0}\selectfont\textcolor[rgb]{0.71,0.12,0.18}{(+15.2)}} \\
\midrule
AI2D & Vanilla & 57.2 & 50.6 & 70.0 & 70.0 & 53.0 & 71.0 & 28.0 & 59.0 & 47.4 & 64.8 \\
 & \vpr & 63.0\,{\fontsize{6.5}{7.0}\selectfont\textcolor[rgb]{0.71,0.12,0.18}{(+5.8)}} & 58.8\,{\fontsize{6.5}{7.0}\selectfont\textcolor[rgb]{0.71,0.12,0.18}{(+8.2)}} & 66.0\,{\fontsize{6.5}{7.0}\selectfont\textcolor[rgb]{0.13,0.52,0.28}{(-4.0)}} & 72.0\,{\fontsize{6.5}{7.0}\selectfont\textcolor[rgb]{0.71,0.12,0.18}{(+2.0)}} & 51.0\,{\fontsize{6.5}{7.0}\selectfont\textcolor[rgb]{0.13,0.52,0.28}{(-2.0)}} & 73.0\,{\fontsize{6.5}{7.0}\selectfont\textcolor[rgb]{0.71,0.12,0.18}{(+2.0)}} & 35.0\,{\fontsize{6.5}{7.0}\selectfont\textcolor[rgb]{0.71,0.12,0.18}{(+7.0)}} & 64.0\,{\fontsize{6.5}{7.0}\selectfont\textcolor[rgb]{0.71,0.12,0.18}{(+5.0)}} & 52.2\,{\fontsize{6.5}{7.0}\selectfont\textcolor[rgb]{0.71,0.12,0.18}{(+4.8)}} & 75.2\,{\fontsize{6.5}{7.0}\selectfont\textcolor[rgb]{0.71,0.12,0.18}{(+10.4)}} \\
\midrule
MMStar & Vanilla & 37.6 & 36.2 & 40.6 & 54.6 & 57.0 & 60.2 & 30.6 & 43.6 & 27.0 & 40.6 \\
 & \vpr & 52.8\,{\fontsize{6.5}{7.0}\selectfont\textcolor[rgb]{0.71,0.12,0.18}{(+15.2)}} & 46.0\,{\fontsize{6.5}{7.0}\selectfont\textcolor[rgb]{0.71,0.12,0.18}{(+9.8)}} & 50.8\,{\fontsize{6.5}{7.0}\selectfont\textcolor[rgb]{0.71,0.12,0.18}{(+10.2)}} & 62.0\,{\fontsize{6.5}{7.0}\selectfont\textcolor[rgb]{0.71,0.12,0.18}{(+7.4)}} & 66.0\,{\fontsize{6.5}{7.0}\selectfont\textcolor[rgb]{0.71,0.12,0.18}{(+9.0)}} & 67.0\,{\fontsize{6.5}{7.0}\selectfont\textcolor[rgb]{0.71,0.12,0.18}{(+6.8)}} & 46.4\,{\fontsize{6.5}{7.0}\selectfont\textcolor[rgb]{0.71,0.12,0.18}{(+15.8)}} & 57.4\,{\fontsize{6.5}{7.0}\selectfont\textcolor[rgb]{0.71,0.12,0.18}{(+13.8)}} & 37.4\,{\fontsize{6.5}{7.0}\selectfont\textcolor[rgb]{0.71,0.12,0.18}{(+10.4)}} & 58.4\,{\fontsize{6.5}{7.0}\selectfont\textcolor[rgb]{0.71,0.12,0.18}{(+17.8)}} \\
\midrule
TextVQA & Vanilla & 37.6 & 43.8 & 60.2 & 65.8 & 70.8 & 70.0 & 37.2 & 51.8 & 29.8 & 61.4 \\
 & \vpr & 71.6\,{\fontsize{6.5}{7.0}\selectfont\textcolor[rgb]{0.71,0.12,0.18}{(+34.0)}} & 62.2\,{\fontsize{6.5}{7.0}\selectfont\textcolor[rgb]{0.71,0.12,0.18}{(+18.4)}} & 65.6\,{\fontsize{6.5}{7.0}\selectfont\textcolor[rgb]{0.71,0.12,0.18}{(+5.4)}} & 69.6\,{\fontsize{6.5}{7.0}\selectfont\textcolor[rgb]{0.71,0.12,0.18}{(+3.8)}} & 74.8\,{\fontsize{6.5}{7.0}\selectfont\textcolor[rgb]{0.71,0.12,0.18}{(+4.0)}} & 74.8\,{\fontsize{6.5}{7.0}\selectfont\textcolor[rgb]{0.71,0.12,0.18}{(+4.8)}} & 53.2\,{\fontsize{6.5}{7.0}\selectfont\textcolor[rgb]{0.71,0.12,0.18}{(+16.0)}} & 68.0\,{\fontsize{6.5}{7.0}\selectfont\textcolor[rgb]{0.71,0.12,0.18}{(+16.2)}} & 55.4\,{\fontsize{6.5}{7.0}\selectfont\textcolor[rgb]{0.71,0.12,0.18}{(+25.6)}} & 68.0\,{\fontsize{6.5}{7.0}\selectfont\textcolor[rgb]{0.71,0.12,0.18}{(+6.6)}} \\
\midrule
RealWorldQA & Vanilla & 39.0 & 42.2 & 51.6 & 59.2 & 61.2 & 59.0 & 31.2 & 52.0 & 39.0 & 51.2 \\
 & \vpr & 55.8\,{\fontsize{6.5}{7.0}\selectfont\textcolor[rgb]{0.71,0.12,0.18}{(+16.8)}} & 55.8\,{\fontsize{6.5}{7.0}\selectfont\textcolor[rgb]{0.71,0.12,0.18}{(+13.6)}} & 60.0\,{\fontsize{6.5}{7.0}\selectfont\textcolor[rgb]{0.71,0.12,0.18}{(+8.4)}} & 66.8\,{\fontsize{6.5}{7.0}\selectfont\textcolor[rgb]{0.71,0.12,0.18}{(+7.6)}} & 68.8\,{\fontsize{6.5}{7.0}\selectfont\textcolor[rgb]{0.71,0.12,0.18}{(+7.6)}} & 64.0\,{\fontsize{6.5}{7.0}\selectfont\textcolor[rgb]{0.71,0.12,0.18}{(+5.0)}} & 43.4\,{\fontsize{6.5}{7.0}\selectfont\textcolor[rgb]{0.71,0.12,0.18}{(+12.2)}} & 61.4\,{\fontsize{6.5}{7.0}\selectfont\textcolor[rgb]{0.71,0.12,0.18}{(+9.4)}} & 56.4\,{\fontsize{6.5}{7.0}\selectfont\textcolor[rgb]{0.71,0.12,0.18}{(+17.4)}} & 63.8\,{\fontsize{6.5}{7.0}\selectfont\textcolor[rgb]{0.71,0.12,0.18}{(+12.6)}} \\
\bottomrule\end{tabular*}
\end{table*}

\subsection{Execution schedules and native rotary conventions}
We prefill the image and context, apply the offset, then prefill the question and assistant-generation suffix; Vanilla shares this boundary. Let $m_0$ be the position of the query predicting the first answer token. At decoding step $j\geq0$, the query advances to $m_0+j$. The schedules are
\begin{equation}
\delta_j^{\mathrm{static}}=\rho\Delta,\qquad
\delta_j^{\mathrm{cont}}=\rho(\Delta+j).
\label{eq:dynamic}
\end{equation}
Static VPR fixes the offset, allowing image--query separation to grow. Continuous VPR tracks the query, maintaining first-answer separation at $\rho=1$. Both share the pre-question intervention. Updates use saved original visual keys to avoid accumulated rotation error. Static VPR is the default, detailed in Algorithm~\ref{alg:static}.

For 1D RoPE, we preserve native rotary pairing and frequency scaling, including Llama-3.1 scaling. MRoPE maps $(t_i,h_i,w_i)$ to $(t_i+\delta,h_i+\delta,w_i+\delta)$ with native axis-to-dimension assignments, including Qwen3-VL's interleaving rather than replacing it with contiguous blocks.

Dynamic-frequency RoPE requires the cached key's original basis:
\begin{equation}
\widetilde k_i=R_{\mathrm{target}}(p_i+\delta)
R_{\mathrm{cached}}(p_i)^{-1}k_i,\quad\delta\ne0.
\label{eq:native}
\end{equation}
At $\delta=0$, the native cache remains unmodified. For InternVL3, changing frequencies during decoding may require re-expressing even a fixed nonzero offset. This adapts the basis while preserving displacement, unlike continuous VPR's increasing displacement. With fixed frequencies, Equation~\ref{eq:native} reduces to Equation~\ref{eq:shift}, requiring one cache translation.

\subsection{Cost and connection to query anchoring}
\label{sec:query_anchoring}
One translation costs $O(N_\ell H_{kv}N_vd_h)$ for $N_\ell$ layers, $H_{kv}$ KV heads and $N_v$ visual tokens, independent of text length once the cache exists. Frequency adaptation may require further translations. Phase coefficients and saved visual keys require temporary storage. After rephasing, ordinary attention consumes the updated keys directly. \vpr\ supports KV caching and the native attention backend, without attention-map extraction, additional tokens or extra forward passes. Section~\ref{sec:efficiency} measures runtime.

Appendix~\ref{app:foundations} provides the full derivation. At fixed content vectors and frequencies, shifting a visual key to $n+\delta$ or its cross-modal query to $m-\delta$ produces the same relative phase in Equation~\ref{eq:rope}. This links \vpr\ to DIPE-style anchoring. Our static implementation chooses the offset from the input gap and preserves normal query progression; Section~\ref{sec:schedule} tests a continuously updated offset.

\section{Experiments}
\label{sec:experiments}
\subsection{Setup}
\label{sec:setup}
\noindent\mbox{\textbf{Models and tasks.}}\enspace
Table~\ref{tab:main} covers four model families and ten checkpoints: Qwen2.5-VL 3B/7B~\citep{bai2025qwen25vl}, Qwen3-VL 2B/4B/8B-Instruct and 8B-Thinking~\citep{bai2025qwen3vl}, InternVL3 2B/8B~\citep{zhu2025internvl3}, LLaVA-Llama3.1-8B~\citep{checkpoint_llava31_8b,grattafiori2024llama3}, and LLaVA-OneVision-Qwen2-7B~\citep{li2024onevision}. All ten cover every main-table task; main-text diagnostics use InternVL3-2B. We freeze weights and retain native encodings: MRoPE for Qwen2.5, interleaved MRoPE for Qwen3, and one-dimensional RoPE for InternVL3 and LLaVA. InternVL3 uses dynamic NTK factor two; LLaVA-Llama3.1 retains native frequency scaling. OneVision's full input exceeds its unscaled 32K base, an extrapolation setting.

POPE~\citep{li2023pope} and CountBench~\citep{paiss2023countbench} assess object perception and counting; DocVQA~\citep{mathew2020docvqa} and TextVQA~\citep{singh2019textvqa} assess document and scene-text reading; ChartQA~\citep{masry2022chartqa} and AI2D~\citep{kembhavi2016ai2d} assess chart and diagram understanding. MathVista~\citep{lu2023mathvista}, GQA~\citep{hudson2019gqa}, MMStar~\citep{chen2024mmstar}, and RealWorldQA~\citep{xai2024realworldqa} cover mathematical reasoning and broader visual question answering.

\par\addvspace{4pt}
\noindent\mbox{\textbf{Context and controls.}}\enspace
We construct $C_L$ from a fixed corpus of Paul Graham essays provided with DIPE's evaluation resources~\citep{chen2026dipe}, using each model's tokenizer. The essays supply no answer, leaving target evidence in the image. The main setting uses $[I,C_L,Q]$ with 32,768 text tokens, in addition to native image, question and template tokens. The short-context reference removes $C_L$. Static \vpr\ defaults to $\rho=1$. Vanilla and VPR share preprocessing, prompts, prefill boundaries and output budgets.

\par\addvspace{4pt}
\noindent\mbox{\textbf{Inference configuration.}}\enspace
Images are converted to RGB and resized within $1024\times1024$ while preserving aspect ratio, followed by native visual processing. Qwen uses pixel bounds of $256^2$--$1024^2$; InternVL uses at most four patches. Native templates retain task questions and answer options. Inference uses H200 GPUs, BF16 and greedy decoding, stopping at EOS or a cap of 1,024 output tokens for Qwen2.5, Qwen3-Instruct and LLaVA, and 4,096 for Qwen3-Thinking and InternVL. Output caps are independent of input length. Ordinary inference uses SDPA; attention measurements use explicit attention.

\par\addvspace{4pt}
\noindent\mbox{\textbf{Metrics and evaluation sets.}}\enspace
We report ANLS$\times100$ for DocVQA~\citep{mathew2020docvqa} and ten-answer consensus$\times100$ for TextVQA~\citep{singh2019textvqa}, without OCR hints. The remaining tasks use their respective accuracy protocols. ChartQA uses relaxed numeric matching, while multiple-choice tasks score extracted options. Gains are absolute score-point differences. The main comparison uses 500 questions per checkpoint and task, identical for Vanilla and VPR, retaining all questions in scoring; Appendix~\ref{app:settings} provides model configurations, prompts, sampling and scoring details.

\subsection{Recovery across tasks and checkpoints}
Static \vpr\ improves 97 of the 100 model--task scores in Table~\ref{tab:main}. All ten checkpoints improve on TextVQA, MMStar, and RealWorldQA. The three decreases occur on GQA and AI2D. The improvements span perception, reading and reasoning, with substantial gains on tasks that require fine visual detail. Qwen2.5-7B gains 24.2 ANLS points on DocVQA, and Qwen3-8B-Instruct gains 18.0 percentage points on ChartQA.

\begin{figure}[!t]\centering
\captionsetup{skip=3pt}
\includegraphics[width=\columnwidth]{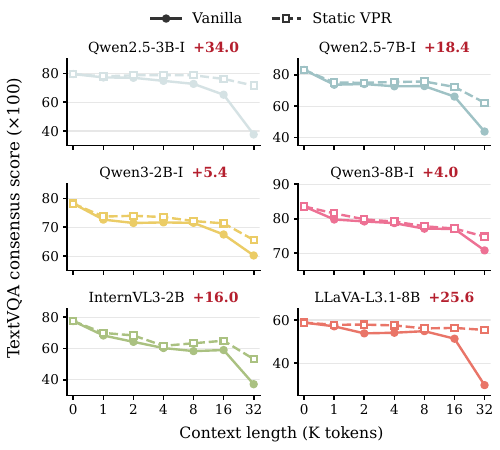}
\caption{TextVQA scores as the passage between the image and question grows. Static VPR scores higher than Vanilla at every nonzero length shown. Each panel uses its own vertical scale.}
\label{fig:textvqa_length}\end{figure}

\subsection{Recovery across context lengths}
\label{sec:length_recovery}
Figure~\ref{fig:textvqa_length} follows identical TextVQA questions across seven lengths. Static VPR improves all 36 nonzero-length comparisons. At 32K, Qwen2.5-3B rises from 37.6 to 71.6, recovering 34.0 of the 42.0 points lost from its 79.6 short-context score. All six models nevertheless remain below their short-context references, indicating substantial but incomplete recovery. Appendix Figure~\ref{fig:length_recovery} reports POPE.

\begin{figure}[!t]\centering
\includegraphics[width=\columnwidth]{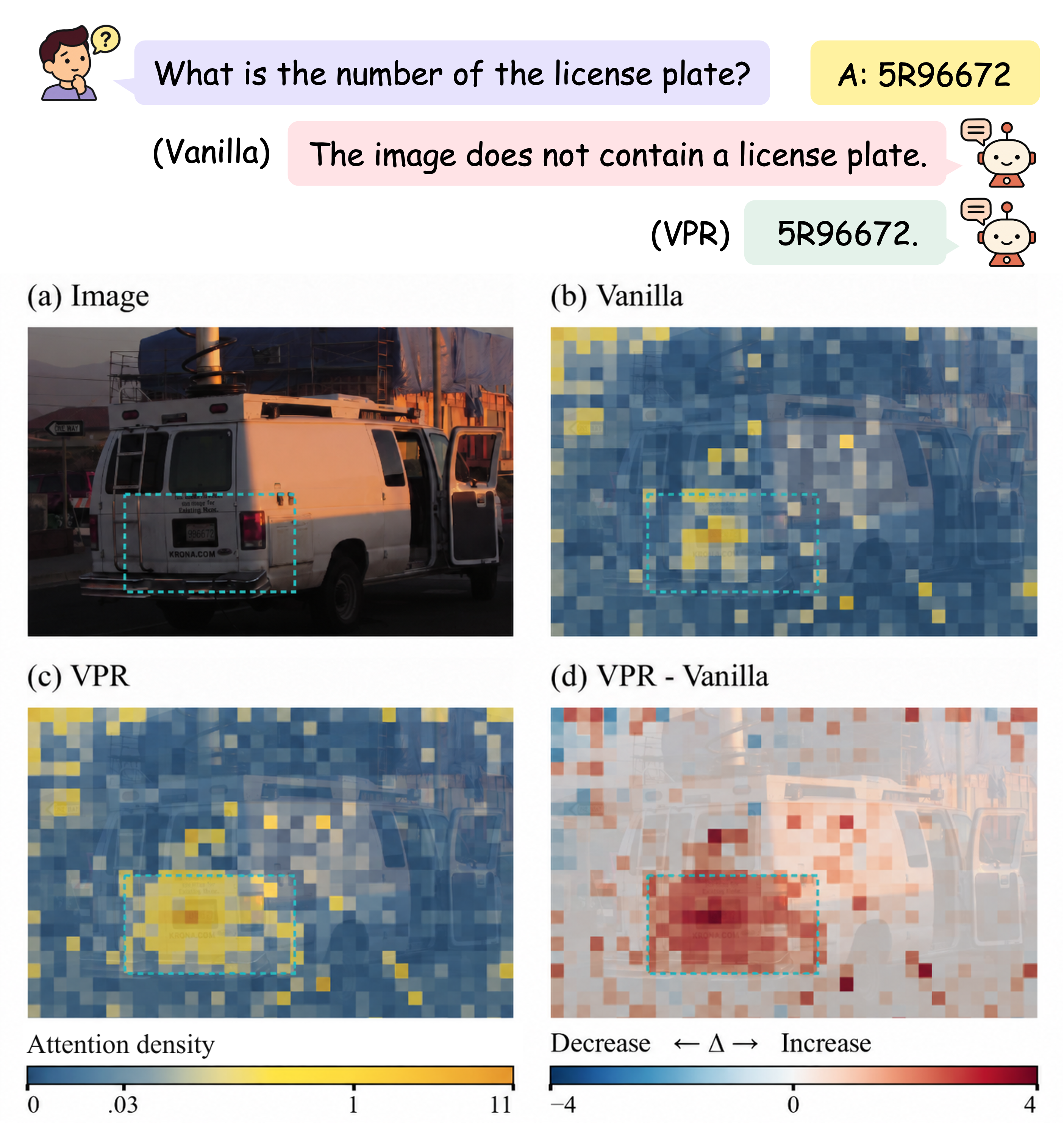}
\caption{VPR recovers the license-plate answer. Dashed boxes highlight the plate region; attention maps share a continuous scale. Red and blue in the difference map indicate increases and decreases, respectively.}
\label{fig:visual_example}\end{figure}

\subsection{Recovery across image positions}
\label{sec:layout_recovery}
Redistributing the same 32K passage in Figure~\ref{fig:layout_scan}, VPR improves all 24 layouts with text after the image; before-only layouts require no compensation. Qwen2.5-3B stays within 82.2--84.6\%, while Vanilla declines as more text follows the image. Recovery remains incomplete for Qwen2.5-7B at 32K-after. InternVL3-2B's before-only score of 70.6\% also falls below its 79.6\% short reference. Re-anchoring improves positional access while retaining representations shaped by the long prefix.

\subsection{Changes in visual attention}
\label{sec:attention_results}
Figure~\ref{fig:visual_example} shows a TextVQA correction at 32K. Qwen2.5-VL-7B-Instruct denies that the image contains a license plate; static \vpr\ returns \texttt{5R96672} correctly, with stronger attention around the plate.

\begin{figure}[!tbp]\centering
\captionsetup{skip=2pt}
\includegraphics[width=\columnwidth]{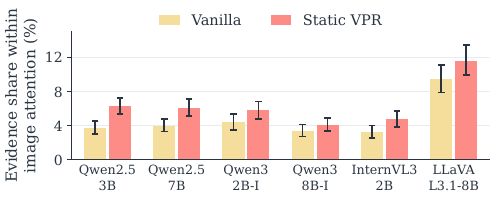}
\caption{Attention share on answer-bearing text within the image. Bars show means with 95\% bootstrap intervals.}
\label{fig:evidence_regions}
\vspace{-3pt}
\end{figure}

Figure~\ref{fig:evidence_regions} tests this redistribution across images by measuring the share of visual attention assigned to text containing the reference answer. Static \vpr\ raises the mean share in all six checkpoints by 0.70--2.57 percentage points. The share rises on 88--95\% of images per checkpoint, and every paired 95\% bootstrap interval for the mean gain lies above zero. Normalizing by total image attention distinguishes redistribution toward answer-bearing text from an overall increase in visual attention. This supports more targeted access to visual evidence. Appendix~\ref{app:evidence_regions} defines the measurement and spatial mapping.

At fixed queries and keys, uniform image-logit amplification preserves the within-image distribution; rephasing can change it.

\subsection{Choosing the compensation strength}
\label{sec:strength}
Table~\ref{tab:strength} separates the reference range $0\leq\rho\leq1$ from beyond-reference shifts. The default $\rho=1$ achieves each checkpoint's highest accuracy and the highest mean, 77.4\%. Qwen3-2B reaches 84.0\%, versus 83.0\% at $\rho=0.75$. This supports restoring the short-context separation at question onset.

\begin{table}[!tbp]\centering
\caption{Long-context POPE accuracy across compensation strengths. Shading marks the default $\rho=1$. Bold marks row maxima. All six checkpoints contribute equally to the mean.}
\label{tab:strength}
\begin{tikzpicture}[x=.01\columnwidth,y=-1pt,
  every node/.style={inner sep=0pt,outer sep=0pt,font=\fontsize{7.1}{8.5}\selectfont}]
\path[use as bounding box] (0,0) rectangle (100,97);
\fill[cvprblue!13] (65.15,12.2) rectangle (74.85,97);
\draw[line width=.7pt] (0,0) -- (100,0);
\node at (50,6.0) {Reference range};
\node[text=black!70] at (90.4,6.0) {Beyond reference};
\draw[line width=.32pt] (25.5,12) -- (74.85,12);
\draw[line width=.32pt] (79.8,12) -- (100,12);
\node[anchor=west] at (0,18) {Model / $\rho$};
\node at (30,18) {0};
\node at (40,18) {.25};
\node at (50,18) {.5};
\node at (60,18) {.75};
\node at (70,18) {\textcolor{cvprblue}{\textbf{1}}};
\node at (85,18) {1.25};
\node at (96,18) {1.5};
\draw[line width=.4pt] (0,24) -- (100,24);
\node[anchor=west] at (0,30) {Qwen2.5-3B-I};
\node at (30,30) {64.6};
\node at (40,30) {67.4};
\node at (50,30) {72.8};
\node at (60,30) {75.8};
\node at (70,30) {\textbf{82.2}};
\node at (85,30) {75.4};
\node at (96,30) {64.2};
\node[anchor=west] at (0,40) {Qwen2.5-7B-I};
\node at (30,40) {51.6};
\node at (40,40) {51.2};
\node at (50,40) {51.4};
\node at (60,40) {53.6};
\node at (70,40) {\textbf{63.4}};
\node at (85,40) {52.2};
\node at (96,40) {44.8};
\node[anchor=west] at (0,50) {Qwen3-2B-I};
\node at (30,50) {75.4};
\node at (40,50) {80.0};
\node at (50,50) {81.0};
\node at (60,50) {83.0};
\node at (70,50) {\textbf{84.0}};
\node at (85,50) {75.0};
\node at (96,50) {72.0};
\node[anchor=west] at (0,60) {Qwen3-8B-I};
\node at (30,60) {77.8};
\node at (40,60) {80.0};
\node at (50,60) {82.0};
\node at (60,60) {84.0};
\node at (70,60) {\textbf{87.0}};
\node at (85,60) {83.0};
\node at (96,60) {80.0};
\node[anchor=west] at (0,70) {InternVL3-2B};
\node at (30,70) {50.0};
\node at (40,70) {51.0};
\node at (50,70) {52.0};
\node at (60,70) {50.0};
\node at (70,70) {\textbf{63.0}};
\node at (85,70) {46.0};
\node at (96,70) {16.0};
\node[anchor=west] at (0,80) {LLaVA-L3.1-8B};
\node at (30,80) {58.2};
\node at (40,80) {67.0};
\node at (50,80) {74.2};
\node at (60,80) {81.0};
\node at (70,80) {\textbf{84.6}};
\node at (85,80) {68.4};
\node at (96,80) {50.0};
\draw[line width=.32pt] (0,85.75) -- (100,85.75);
\node[anchor=west] at (0,91.5) {\textbf{Mean}};
\node at (30,91.5) {62.9};
\node at (40,91.5) {66.1};
\node at (50,91.5) {68.9};
\node at (60,91.5) {71.2};
\node at (70,91.5) {\textbf{77.4}};
\node at (85,91.5) {66.7};
\node at (96,91.5) {54.5};
\draw[line width=.7pt] (0,97) -- (100,97);
\end{tikzpicture}
\end{table}

Beyond-reference shifts reduce accuracy: at $\rho=1.5$, LLaVA falls to 50.0\% and InternVL3-2B to 16.0\%. These settings can reverse effective image--query ordering and involve positional extrapolation; they explore beyond-reference behavior rather than isolate one cause of failure.

\subsection{Update schedule and execution cost}
\label{sec:schedule}
Table~\ref{tab:schedule} compares static and continuous VPR at $\rho=1$ using the schedules in Equation~\ref{eq:dynamic}. Continuous scores differ from static by $-1.2$ to $+1.1$ points on TextVQA and $-2.8$ to $+2.0$ on MathVista. Tracking the query therefore provides no consistent accuracy advantage.
\begin{table}[!htbp]\centering\fontsize{8}{9.6}\selectfont
\setlength{\tabcolsep}{0pt}
\caption{Static and continuous VPR. Diff. denotes Continuous minus Static.}\label{tab:schedule}
\begin{tabular*}{\columnwidth}{@{\extracolsep{\fill}}llrrrr@{}}\toprule Task & Model & Vanilla & Static & Cont. & Diff. \\\midrule
TextVQA & Qwen2.5-7B & 43.8 & 62.2 & 61.8 & -0.4 \\
 & Qwen3-8B-I & 70.8 & 74.8 & 75.0 & +0.2 \\
 & InternVL3-2B & 37.2 & 53.2 & 54.3 & +1.1 \\
 & LLaVA-L3.1-8B & 29.8 & 55.4 & 54.2 & -1.2 \\
\midrule
MathVista & Qwen2.5-7B & 38.4 & 47.4 & 49.4 & +2.0 \\
 & Qwen3-8B-I & 61.0 & 63.0 & 60.2 & -2.8 \\
 & InternVL3-2B & 38.8 & 44.6 & 46.6 & +2.0 \\
 & LLaVA-L3.1-8B & 23.8 & 26.8 & 27.0 & +0.2 \\
\bottomrule\end{tabular*}\end{table}

\label{sec:efficiency}
Table~\ref{tab:efficiency} measures split prefill plus 128 or 512 generated tokens for four checkpoints on the same hardware. Disabling EOS stopping fixes the workload, separating cache-update cost from answer-length differences. Qwen and LLaVA retain fixed frequencies: static VPR updates keys once, adding 0.06--0.33\% at 32K; continuous updates add 6.23--10.91\%. Against native single-pass prefill, static overhead at 128 output tokens is 2.29--3.50\%, including split-prefill cost.
\begin{table}[!htbp]\centering\fontsize{7.8}{9.2}\selectfont
\vspace{-7pt}
\setlength{\tabcolsep}{1.4pt}\renewcommand{\arraystretch}{0.95}
\caption{Inference time and overhead relative to Vanilla with matched split prefill.}\label{tab:efficiency}
\begin{tabular*}{\columnwidth}{@{\extracolsep{\fill}}llrrrrr@{}}\toprule
 & & \multicolumn{1}{c}{Vanilla} & \multicolumn{2}{c}{Static VPR} & \multicolumn{2}{c}{Continuous VPR} \\
\cmidrule(lr){3-3}\cmidrule(lr){4-5}\cmidrule(l){6-7}
Output & Model & Time (s) & Time (s) & Extra (\%) & Time (s) & Extra (\%) \\\midrule
128 & Qwen2.5-7B & 3.465 & 3.470 & +0.13 & 3.709 & +7.03 \\
 & Qwen3-8B-I & 4.738 & 4.741 & +0.06 & 5.033 & +6.23 \\
 & InternVL3-2B & 2.283 & 2.934 & +28.51 & 2.941 & +28.82 \\
 & LLaVA-L3.1-8B & 3.991 & 3.997 & +0.14 & 4.251 & +6.52 \\
\midrule
512 & Qwen2.5-7B & 9.416 & 9.447 & +0.33 & 10.443 & +10.91 \\
 & Qwen3-8B-I & 12.940 & 12.969 & +0.22 & 14.121 & +9.12 \\
 & InternVL3-2B & 7.484 & 10.015 & +33.82 & 10.019 & +33.87 \\
 & LLaVA-L3.1-8B & 10.011 & 10.023 & +0.12 & 11.051 & +10.39 \\
\bottomrule\end{tabular*}\end{table}

For InternVL3-2B, the complete input exceeds its 32K base, activating dynamic NTK scaling. Changing frequencies require rephasing even with a fixed offset through Equation~\ref{eq:native}. At 512 output tokens, both schedules make 513 updates: static takes 10.015\,s and continuous 10.019\,s. Their similar runtimes reflect shared frequency-adaptation work. At 8K, frequencies remain fixed: static makes one update (+0.04\%), versus 512 (+36.71\%) for continuous VPR. Appendix~\ref{app:efficiency} reports the full measurements. These results favor static VPR, especially with fixed native frequencies.

\subsection{Comparison with attention interventions}
\label{sec:attention_baselines}
We evaluate four training-free attention interventions on the same Qwen2.5-VL-7B-Instruct checkpoint and POPE adversarial questions: VAF~\citep{yin2025clearsight}, VGA~\citep{zhao2026vga}, VAR~\citep{kang2025var}, and VRGA~\citep{peng2026vrga}. At 0K, VPR applies zero displacement and matches Vanilla at 85.8\%. Table~\ref{tab:attention_baselines} shows that VPR achieves the highest accuracy at every tested nonzero context length, gaining 5.0--12.0 percentage points over Vanilla. VAF is the closest competitor: at 32K, VPR reaches 63.4\% and VAF reaches 62.0\%, compared with 51.6\% for Vanilla.

\begin{table}[!htbp]\centering
\caption{Training-free interventions on Qwen2.5-VL-7B-Instruct across POPE context lengths. Cost reports inference time and overhead relative to Vanilla.}\label{tab:attention_baselines}
\fontsize{7.5}{9}\selectfont\setlength{\tabcolsep}{1.4pt}
\begin{tabular*}{\columnwidth}{@{\extracolsep{\fill}}lrrrrrrr@{}}\toprule
 & \multicolumn{5}{c}{POPE accuracy (\%)} & \multicolumn{2}{c}{Cost} \\
\cmidrule(lr){2-6}\cmidrule(l){7-8}
Method & 0K & 8K & 16K & 24K & 32K & Time (s) & Extra (\%) \\\midrule
Vanilla & \textcolor[rgb]{0.10,0.32,0.64}{\underline{85.8}} & 72.8 & 63.0 & 56.4 & 51.6 & 3.465 & ref. \\
\vpr & \textcolor[rgb]{0.10,0.32,0.64}{\underline{85.8}} & \textcolor[rgb]{0.70,0.10,0.16}{\textbf{77.8}} & \textcolor[rgb]{0.70,0.10,0.16}{\textbf{74.6}} & \textcolor[rgb]{0.70,0.10,0.16}{\textbf{68.4}} & \textcolor[rgb]{0.70,0.10,0.16}{\textbf{63.4}} & \textcolor[rgb]{0.70,0.10,0.16}{\textbf{3.470}} & \textcolor[rgb]{0.70,0.10,0.16}{\textbf{+0.13}} \\
VAF & \textcolor[rgb]{0.70,0.10,0.16}{\textbf{86.0}} & \textcolor[rgb]{0.10,0.32,0.64}{\underline{76.6}} & \textcolor[rgb]{0.10,0.32,0.64}{\underline{73.6}} & \textcolor[rgb]{0.10,0.32,0.64}{\underline{67.2}} & \textcolor[rgb]{0.10,0.32,0.64}{\underline{62.0}} & \textcolor[rgb]{0.10,0.32,0.64}{\underline{3.631}} & \textcolor[rgb]{0.10,0.32,0.64}{\underline{+4.8}} \\
VGA & \textcolor[rgb]{0.10,0.32,0.64}{\underline{85.8}} & 72.8 & 63.0 & 56.4 & 52.8 & 4.081 & +17.8 \\
VAR & 85.4 & 73.4 & 63.4 & 56.8 & 54.2 & 19.408 & +460.1 \\
VRGA & 85.2 & 72.8 & 62.0 & 56.4 & 52.0 & 6.489 & +87.3 \\
\bottomrule\end{tabular*}
\end{table}

For the cost comparison, we fix the input passage at 32K and generate 128 tokens on an H200. Static VPR takes 3.470\,s per request and adds 0.13\% over Vanilla, the smallest relative overhead among the five interventions. VAF adds 4.8\%. On this checkpoint, a single update to the cached image keys therefore recovers long-context accuracy with a measured cost close to Vanilla.

\subsection{Joint use of image and text evidence}
\label{sec:joint_mapping}
We pair TextVQA images with text mappings from visual details to four-letter answer codes. The model must identify the relevant image detail and apply the supplied mapping, making visual access a prerequisite for the final answer. Re-anchoring adjusts the effective image--question positional relation, helping the query access visual evidence while retaining the intervening text for the mapping step.

Table~\ref{tab:joint_mapping} shows gains of 3.50--7.00 percentage points across both checkpoints and all four context lengths. The benefit extends beyond direct visual answering to a task requiring a text lookup. By changing visual key phases while retaining the text, VPR improves access to details needed for the mapping step. Qwen2.5's low absolute accuracy nevertheless leaves substantial room for improvement. Appendix~\ref{app:joint_mapping} details the protocol and blank-image control.

\begin{table}[!t]
\centering
\caption{Joint image and text lookup across context lengths. Scores measure final answer code accuracy; $\Delta$ is the static VPR gain in percentage points.}
\label{tab:joint_mapping}
\fontsize{7.5}{9}\selectfont
\setlength{\tabcolsep}{2pt}
\renewcommand{\arraystretch}{1.13}
\begin{tabular*}{\columnwidth}{@{\extracolsep{\fill}}llrrrr@{}}
\toprule
& & \multicolumn{4}{c}{Joint lookup accuracy (\%)} \\
\cmidrule(l){3-6}
Model & Method & 8K & 16K & 24K & 32K \\
\midrule
Qwen2.5-7B & Vanilla & 18.75 & 14.25 & 11.75 & 11.00 \\
 & VPR & \textbf{23.25} & \textbf{21.25} & \textbf{18.00} & \textbf{17.00} \\
 & $\Delta$ & \textcolor[rgb]{0.70,0.10,0.16}{+4.50} & \textcolor[rgb]{0.70,0.10,0.16}{+7.00} & \textcolor[rgb]{0.70,0.10,0.16}{+6.25} & \textcolor[rgb]{0.70,0.10,0.16}{+6.00} \\
\midrule
Qwen3-8B-I & Vanilla & 72.25 & 64.25 & 60.50 & 65.75 \\
 & VPR & \textbf{77.50} & \textbf{70.25} & \textbf{64.50} & \textbf{69.25} \\
 & $\Delta$ & \textcolor[rgb]{0.70,0.10,0.16}{+5.25} & \textcolor[rgb]{0.70,0.10,0.16}{+6.00} & \textcolor[rgb]{0.70,0.10,0.16}{+4.00} & \textcolor[rgb]{0.70,0.10,0.16}{+3.50} \\
\bottomrule
\end{tabular*}
\end{table}

\section{Discussion}
\label{sec:discussion}
Our results identify positional relations as an actionable factor in long-context visual access. Position-only controls change performance with input tokens fixed, while VPR recovers accuracy without removing competing text keys. The joint lookup results in Section~\ref{sec:joint_mapping} further suggest that improved visual access can benefit answers requiring textual evidence. Position alone does not explain every loss: InternVL3-2B remains below its short-context reference even when the added passage precedes the image. Extending re-anchoring to evidence distributed across multiple images and documents, and revisited throughout generated reasoning, is a natural next step.

\section{Conclusion}
We identify positional relations as an actionable factor in long-context visual access and introduce Visual Position Re-Anchoring (VPR), a training-free intervention that rephases cached visual keys while preserving standard attention computation. Across ten checkpoints and ten benchmarks, static VPR improves 97 of 100 model--task scores. Evidence-region analysis shows increased attention to answer-bearing content, while joint lookup experiments extend the benefit to answers requiring both image and text evidence. With fixed rotary frequencies, a single cache update achieves these gains with minimal runtime overhead. Together, these findings establish cache-level positional intervention as a practical approach to sustaining visual access in long-context MLLMs.

\FloatBarrier
\label{page:main_end}

{\small
\bibliographystyle{ieeenat_fullname}
\bibliography{references}
}
\clearpage
\clearcaptionsetup[skip]{table}
\onecolumn
\appendix
\phantomsection\label{page:supp_start}
\begin{center}
{\Large\bfseries Keep Images in Sight: Visual Position Re-Anchoring\\for Long-Context MLLMs\par}
\medskip
{\large Supplementary Material\par}
\end{center}
\section{Positional Foundations and Derivations}
\label{app:foundations}
\begingroup
\setlength{\abovedisplayskip}{6pt plus 2pt minus 1pt}
\setlength{\belowdisplayskip}{6pt plus 2pt minus 1pt}
\setlength{\abovedisplayshortskip}{3pt plus 1pt}
\setlength{\belowdisplayshortskip}{4pt plus 1pt}
\subsection{Notation and derivation scope}
\label{app:derivation}
We derive Section~\ref{sec:method}'s positional operation for one decoder layer and head, in exact arithmetic unless stated otherwise. The operation applies to every visual KV head at every layer.

Let $\bar q_m,\bar k_i\in\R^{d_h}$ be pre-RoPE content vectors at native positions $m,p_i$. A local positional comparison holds these vectors fixed; it does not equate them with a separate short-context run. Reference gaps use unchanged image geometry and native templates. We first assume fixed frequencies, then extend to multimodal coordinates (Section~\ref{app:multiaxis_derivation}) and changing frequencies (Section~\ref{app:native_derivation}).

\begin{center}
\small
\begin{tabular}{@{}ll@{}}
\toprule
Symbol & Meaning \\
\midrule
$p_i,\ a,\ p_Q$ & Visual-token position, first post-image position, first question position \\
$b_0,\ \Delta$ & Native short-template gap and extra input gap \\
$\rho,\ \delta$ & Dimensionless compensation strength and positional displacement \\
$d_i^{(0)}$ & Signed question--image separation in the short-context reference \\
$m_0,\ j$ & Query position predicting the first answer token and decoding-step index \\
$g_i^{(0)}$ & Short-reference separation at the first-answer query \\
\bottomrule
\end{tabular}
\end{center}

\subsection{From rotary attention to a cache-only translation}
\label{app:rotation_derivation}
\paragraph{Rotation composition.}
For a two-dimensional rotary pair, define
\begin{equation}
U(\phi)=
\begin{pmatrix}\cos\phi&-\sin\phi\\ \sin\phi&\cos\phi\end{pmatrix},
\qquad
\rot(p)=\operatorname{diag}_{r=1}^{d_r/2} U(p\theta_r),
\label{eq:app_rotation}
\end{equation}
where $d_r\leq d_h$ is the rotated dimension and $\theta_r$ is the frequency of pair $r$. Identity blocks on any unrotated dimensions are implicit. The checkpoint's pairing convention is a fixed reordering of these coordinates. Angle-addition identities give
\begin{equation}
U(\alpha)U(\beta)=U(\alpha+\beta),
\qquad U(\alpha)^\top=U(-\alpha)=U(\alpha)^{-1}.
\label{eq:app_composition}
\end{equation}
Consequently, $\rot(u)\rot(v)=\rot(u+v)$ and $\rot(u)^\top=\rot(-u)$, also under the native coordinate reordering.

\paragraph{Relative phase in the logit.}
With $q_m=\rot(m)\bar q_m$ and $k_i=\rot(p_i)\bar k_i$, the unmasked attention logit is
\begin{equation}
\begin{aligned}
s_{m,i}
&=\frac{q_m^\top k_i}{\sqrt{d_h}}
=\frac{\bar q_m^\top\rot(m)^\top\rot(p_i)\bar k_i}{\sqrt{d_h}}\\
&=\frac{\bar q_m^\top\rot(p_i-m)\bar k_i}{\sqrt{d_h}}.
\end{aligned}
\label{eq:app_logit}
\end{equation}
The mask is applied to the original sequence slots. A change of effective rotary position does not change which slots are visible.

\paragraph{Translating an already cached key.}
Applying a displacement $\delta$ directly to the cached vector gives
\begin{equation}
\begin{aligned}
\widetilde k_i
&=\rot(\delta)k_i
=\rot(\delta)\rot(p_i)\bar k_i
=\rot(p_i+\delta)\bar k_i,\\
\widetilde s_{m,i}
&=\frac{\bar q_m^\top\rot(p_i+\delta-m)\bar k_i}{\sqrt{d_h}}.
\end{aligned}
\label{eq:app_cached_translation}
\end{equation}
Thus the operation realizes the key position $p_i+\delta$ without recomputing $\bar k_i$. It modifies visual matching while retaining the stored values and text cache. Later queries can change because they attend to the modified cache.

\paragraph{Relation to query anchoring.}
For the same content vectors and frequencies,
\begin{equation}
\bigl[\rot(m)\bar q_m\bigr]^\top
\bigl[\rot(p_i+\delta)\bar k_i\bigr]
=
\bigl[\rot(m-\delta)\bar q_m\bigr]^\top
\bigl[\rot(p_i)\bar k_i\bigr].
\label{eq:app_query_equivalence}
\end{equation}
Both sides have relative phase $p_i-m+\delta$. This is the local cross-modal equivalence underlying the connection to query anchoring. Equivalence of a complete attention output additionally requires identical logits for the other keys, identical values and identical masks. Shifting a query against text keys as well would change those logits, whereas VPR transforms only visual keys.

\subsection{Reference-gap compensation and the meaning of \texorpdfstring{$\rho$}{rho}}
\label{app:rho_derivation}
\paragraph{Constructing the reference.}
Use superscript $(0)$ for positions in the short-context template. By definition,
\begin{equation}
b_0=p_Q^{(0)}-a^{(0)},\qquad
\Delta=(p_Q-a)-b_0.
\label{eq:app_gap_definition}
\end{equation}
Keeping the image's native geometry unchanged gives $a-p_i=a^{(0)}-p_i^{(0)}$. This permits a common translation of the image and its anchor, as occurs when text is placed before the image. Therefore,
\begin{equation}
\begin{aligned}
d_i=p_Q-p_i
&=(p_Q-a)+(a-p_i)\\
&=(b_0+\Delta)+(a^{(0)}-p_i^{(0)})\\
&=p_Q^{(0)}-p_i^{(0)}+\Delta
=d_i^{(0)}+\Delta.
\end{aligned}
\label{eq:app_gap_expansion}
\end{equation}
This calculation uses native position IDs. The extra position gap $\Delta$ is distinct from the nominal number of inserted text tokens, the full evidence span $d_i$, and the output-token budget.

\paragraph{Applying the strength parameter.}
Setting $\delta=\rho\Delta$ and substituting Equation~\ref{eq:app_gap_expansion} yields
\begin{equation}
\begin{aligned}
\widetilde d_i
&=p_Q-(p_i+\rho\Delta)
=d_i-\rho\Delta\\
&=\boxed{d_i^{(0)}+(1-\rho)\Delta}.
\end{aligned}
\label{eq:app_compensated_gap}
\end{equation}
For $\Delta>0$, $\rho=0$ retains the full added gap, $\rho=\tfrac12$ retains half, $\rho=1$ removes it, and $\rho=\tfrac32$ shifts the key beyond the reference by $\Delta/2$. Overshooting the reference does not necessarily place the key beyond the query: the signed separation becomes negative only when $\rho>1+d_i^{(0)}/\Delta$. The physical cache order remains unchanged in either case.

\paragraph{Why unit strength is the reference choice.}
Equation~\ref{eq:app_compensated_gap} gives the coordinate-restoration error
\begin{equation}
E_{\mathrm{pos}}(\rho)
=\frac{1}{N_v}\sum_{i\in\imgset}
\bigl(\widetilde d_i-d_i^{(0)}\bigr)^2
=(1-\rho)^2\Delta^2.
\label{eq:app_position_error}
\end{equation}
When $\Delta\ne0$, this nonnegative quantity has its unique zero and minimum at $\rho=1$. When $\Delta=0$, every static strength gives zero displacement. $E_{\mathrm{pos}}$ describes exact coordinate restoration; it is not a training loss or an accuracy objective.

The reason this reference is useful is that it restores the positional relation of the ordinary short input while retaining the long-prefix computation. Its effect on answer accuracy is established by the strength experiment, rather than by minimizing $E_{\mathrm{pos}}$. In particular, for one rotary pair with content vectors $(u,v)$ and $(x,y)$ and angle $\phi=\theta_r(p_i+\delta-m)$, its dot-product contribution is
\begin{equation}
(ux+vy)\cos\phi+(vx-uy)\sin\phi.
\label{eq:app_nonmonotonic}
\end{equation}
Its derivative with respect to $\delta$ is $\theta_r[-(ux+vy)\sin\phi+(vx-uy)\cos\phi]$, which has no fixed sign. Softmax also compares the visual logits with all other visible keys. Accordingly, coordinate restoration neither implies monotonic attention growth nor requires every example to attain its best accuracy at unit strength.

\subsection{Static and continuous schedules during generation}
\label{app:schedule_derivation}
Let $\ell_Q=m_0-p_Q$ be the native positional span from the first question token to the query predicting the first answer token, including the generation suffix. This span is fixed for the same question and template. Define $g_i^{(0)}=d_i^{(0)}+\ell_Q$, the corresponding first-answer separation in the short reference. Since $m_0-p_i=g_i^{(0)}+\Delta$, a text decoding query at $m_j=m_0+j$ has separation
\begin{equation}
d_{i,j}=m_j-p_i=g_i^{(0)}+\Delta+j,
\qquad j=0,1,2,\ldots .
\label{eq:app_decode_gap}
\end{equation}
Here $j=0$ is the final prompt query, not the first generated token's own query. The evaluated text decoders advance each applicable position coordinate by one per generated token.

\paragraph{Static schedule.}
The offset $\delta_j^{\mathrm{stat}}=\rho\Delta$ is fixed after context prefill. Subtracting it from Equation~\ref{eq:app_decode_gap} gives
\begin{equation}
d_{i,j}^{\mathrm{stat}}
=\boxed{g_i^{(0)}+(1-\rho)\Delta+j}.
\label{eq:app_static_gap}
\end{equation}
At $\rho=1$, $d_{i,j}^{\mathrm{stat}}=g_i^{(0)}+j$. Static VPR removes the added input gap while retaining ordinary query progression. Its image--query separation grows by one at every decoding step; it matches the short-reference positional trajectory at the same step index, not a constant distance throughout generation.

\paragraph{Continuous schedule.}
The continuously updated offset is $\delta_j^{\mathrm{cont}}=\rho(\Delta+j)$. Hence
\begin{equation}
\begin{aligned}
d_{i,j}^{\mathrm{cont}}
&=g_i^{(0)}+\Delta+j-\rho(\Delta+j)\\
&=\boxed{g_i^{(0)}+(1-\rho)(\Delta+j)}.
\end{aligned}
\label{eq:app_continuous_gap}
\end{equation}
At $\rho=1$, $d_{i,j}^{\mathrm{cont}}=g_i^{(0)}$ for every $j$: the first-answer reference distance is maintained. Both schedules begin with the same offset and diverge by
\begin{equation}
\delta_j^{\mathrm{cont}}-\delta_j^{\mathrm{stat}}=\rho j,
\qquad
d_{i,j}^{\mathrm{stat}}-d_{i,j}^{\mathrm{cont}}=\rho j.
\label{eq:app_schedule_difference}
\end{equation}
Thus $\rho$ controls the fraction of the input gap removed in both schedules, and also the fraction of subsequent query displacement tracked in the continuous schedule. At $\rho=0$ both are native-cache baselines. With $\Delta=0$ and $\rho>0$, static VPR remains the identity, whereas continuous VPR can still act once $j>0$. These positional identities compare equal step indices; different schedules may generate different answer tokens or stop at different steps.

\subsection{Multimodal coordinates and preserved structure}
\label{app:multiaxis_derivation}
Let $\mathbf p_i=(t_i,h_i,w_i)^\top$ be a visual coordinate and let $a(r)\in\{t,h,w\}$ select the native axis assigned to rotary pair $r$. In the checkpoint's pair ordering,
\begin{equation}
R_\Theta(\mathbf p_i)
=\operatorname{diag}_r U(\theta_r p_{i,a(r)}).
\label{eq:app_mrope}
\end{equation}
For $\mathbf 1=(1,1,1)^\top$, a shared translation gives
\begin{equation}
R_\Theta(\mathbf p_i+\delta\mathbf 1)
=T_\delta R_\Theta(\mathbf p_i),
\qquad T_\delta=\operatorname{diag}_r U(\theta_r\delta).
\label{eq:app_mrope_translation}
\end{equation}
The identity holds for contiguous or interleaved axis assignments because each native pair receives its own prescribed frequency. The implementation retains that assignment. When additional text advances each question coordinate by the same $\Delta$, the reference calculation applies componentwise:
\begin{equation}
\widetilde{\mathbf d}_i
=\mathbf d_i^{(0)}+(1-\rho)\Delta\mathbf 1.
\label{eq:app_mrope_gap}
\end{equation}
The scalar-shift derivation requires this common coordinate increment; distinct axis increments would require an axis-specific extension.

Two structural properties now follow. First, within-image coordinate differences are unchanged:
\begin{equation}
(\mathbf p_i+\delta\mathbf 1)-(\mathbf p_k+\delta\mathbf 1)
=\mathbf p_i-\mathbf p_k.
\label{eq:app_coordinate_invariant}
\end{equation}
Second, for visual keys in the same head and fixed frequency basis, $T_\delta^\top T_\delta=I$ implies
\begin{equation}
\|T_\delta k_i\|_2^2=\|k_i\|_2^2,
\qquad
(T_\delta k_i)^\top(T_\delta k_k)=k_i^\top k_k.
\label{eq:app_gram_invariant}
\end{equation}
The common rotation preserves the visual-key Gram matrix while changing compatibility with unshifted queries. This property concerns the cached representations at the intervention, not the hidden states subsequently produced by the model.

\subsection{Native endpoint composition and changing frequencies}
\label{app:native_derivation}
\paragraph{Endpoint coefficients.}
Represent the native source or target transform of one rotary pair as
\begin{equation}
A_b=\begin{pmatrix}c_b&-s_b\\s_b&c_b\end{pmatrix},\qquad b\in\{0,1\}.
\end{equation}
For source coefficients with $c_0^2+s_0^2>0$, direct inversion and multiplication give
\begin{equation}
\begin{aligned}
A_0^{-1}
&=\frac{1}{c_0^2+s_0^2}
\begin{pmatrix}c_0&s_0\\-s_0&c_0\end{pmatrix},\\
A_1A_0^{-1}
&=\begin{pmatrix}c_\Delta&-s_\Delta\\s_\Delta&c_\Delta\end{pmatrix},\\
c_\Delta&=\frac{c_1c_0+s_1s_0}{c_0^2+s_0^2},\qquad
s_\Delta=\frac{s_1c_0-c_1s_0}{c_0^2+s_0^2}.
\end{aligned}
\label{eq:app_endpoint_coefficients}
\end{equation}
For exact unit sine/cosine pairs the denominator is one and the result is the angle-difference rotation. For finite-precision endpoint coefficients, this expression is the exact algebraic composition of the represented matrices before subsequent numerical rounding. It does not undo rounding already introduced when the cached vector was formed or quantized.

The implementation supports incremental and native-endpoint phases. Their exact-arithmetic equivalence does not require separately computed floating-point angles to coincide. Continuous updates use saved original image keys rather than repeatedly rotating the last rounded result. The overshoot, schedule and efficiency evaluations use native endpoint coefficients; the Qwen2.5-7B strength display includes incremental coefficients below unit strength.

\paragraph{Changing frequency bases.}
Write an original cached key as $k_i^{\mathrm{src}}=R_{\mathrm{src}}(p_i)\bar k_i$. At decoding step $j$, the required target is $R_{\mathrm{tgt},j}(p_i+\delta_j)\bar k_i$. Eliminating $\bar k_i$ gives
\begin{equation}
\begin{aligned}
\widetilde k_{i,j}
&=R_{\mathrm{tgt},j}(p_i+\delta_j)
R_{\mathrm{src}}(p_i)^{-1}k_i^{\mathrm{src}},\\
\phi_{i,r,j}
&=(p_i+\delta_j)\theta_{r,j}^{\mathrm{tgt}}
-p_i\theta_r^{\mathrm{src}}\\
&=\delta_j\theta_{r,j}^{\mathrm{tgt}}
+p_i(\theta_{r,j}^{\mathrm{tgt}}-\theta_r^{\mathrm{src}}).
\end{aligned}
\label{eq:app_dynamic_basis}
\end{equation}
The second term vanishes for a fixed basis, recovering the common incremental rotation. Otherwise it depends on $p_i$: fixed displacement does not imply a fixed cache transform. This explains why static VPR may require repeated re-expression for InternVL3. For pure rotary transforms each key norm is preserved in exact arithmetic, but the common-transform argument for preserving the original cross-key Gram matrix no longer applies when the basis changes. Coordinate differences remain preserved by the translation itself.

The implementation tracks source and target frequency states and resets native frequency state between requests. At zero intervention it bypasses Equation~\ref{eq:app_dynamic_basis} and retains the native cache. This convention matters because setting $\delta_j=0$ in a source-to-target basis conversion need not produce the identity when the two bases differ. At nonzero displacement the target key is re-expressed in the current native basis; the positional schedule and the frequency-update schedule are distinct.

\subsection{Operation count}
Each rotary pair requires a constant number of arithmetic operations. Touching all $N_v$ visual keys in $H_{kv}$ KV heads across $N_\ell$ layers therefore costs $O(N_\ell H_{kv}N_vd_r)$, bounded by $O(N_\ell H_{kv}N_vd_h)$, per translation. With fixed frequencies, static VPR requires one translation; maintaining continuous targets for $B$ predicted answer tokens requires at most $B$ translations, including the initial one. Changing frequencies can require additional cache writes even for a fixed offset. Saved original visual keys occupy $O(N_\ell H_{kv}N_vd_h)$ storage, with additional phase workspace determined by the implementation. These are cache operations rather than additional model forward passes; Section~\ref{sec:efficiency} reports their measured runtime and memory effects.
\par\endgroup

\section{Implementation and Evaluation Protocols}
\label{app:implementation}
\label{app:settings}
\subsection{Cache operation and algorithm}
The implementation locates image indices and native position IDs, then splits the prompt before the first question token. Context prefill constructs the prefix cache. VPR changes visual-key entries at every decoder layer and leaves text keys and all values intact. Question prefill then reads this modified cache. The single-image layout ensures that the evaluated queries occur after the visual input.

The native template determines the short-context separation $b_0$. Zero compensation retains the native cache, including its ordinary frequency updates. For nonzero shifts, rotary pairing, scaling and axis assignment must match the checkpoint.
\begin{figure}[htbp]\small\hrule\vspace{5pt}
\refstepcounter{algorithm}\textbf{Algorithm \thealgorithm: Static Visual Position Re-Anchoring}\par
\label{alg:static}
\textbf{Input:} frozen model $f$, image $I$, context $C$, question $Q$, strength $\rho$, output budget $B$.\par
\begin{enumerate}
\item Prepare the native prompt, position IDs and visual indices $\imgset$. Split before $Q$.
\item Compute $\Delta$ from Equation~\ref{eq:delta}; set $\delta=\rho\Delta$.
\item Prefill the image and context to obtain the KV cache.
\item If $\delta\ne0$, rephase visual keys to their target positions; retain the other cache entries.
\item Prefill the question and generation suffix to obtain logits $z$.
\item Select $y=\argmax z$; stop at EOS or budget $B$. Otherwise decode $y$ to update $z$ and the cache, and repeat.
\end{enumerate}
For changing native frequencies, maintain the fixed target offset with Equation~\ref{eq:native}.\par\vspace{5pt}\hrule
\end{figure}

\subsection{Models, context limits, and preprocessing}
Table~\ref{tab:models} lists the ten evaluated checkpoint releases, with links to their model repositories. It separates reported multimodal training context from the archived maximum-position configuration, the evaluation ceiling, and the output budget. Our default 32K denotes added text; the complete input also contains the image, question, and template tokens.

InternVL3 uses dynamic NTK scaling with factor two, a 32,768 base configuration, and a 65,536 evaluation ceiling. OneVision uses unscaled Qwen2 RoPE ($\theta=10^6$). Its evaluation ceiling is 131,072, whereas its base configuration is 32,768; the full 32K-text condition therefore extrapolates beyond that base. Neither a tokenizer limit nor a larger configured ceiling establishes the length used during multimodal training.

Images are converted to RGB and thumbnail-limited to $1024\times1024$, preserving aspect ratio. Qwen additionally uses pixel bounds of $256^2$ to $1024^2$; InternVL uses at most four patches. Native processors determine visual-token expansion and position IDs. Comparisons within each checkpoint share preprocessing, while cross-family comparisons retain their native visual-token counts.
\begingroup
\small\setlength{\tabcolsep}{3pt}\renewcommand{\arraystretch}{1.12}
\begin{table}[htbp]\centering
\caption{Model releases, training context, and evaluation settings. Checkpoint names link to the evaluated releases. Training entries refer to the sourced recipes below; Config. is the archived maximum-position setting, and Eval. ceiling is the configured evaluation ceiling, not the tested input length. Output is the generation cap.}\label{tab:models}\begin{tabular}{@{}llcrrr@{}}
\toprule
Checkpoint & RoPE & Training\textsuperscript{*} & Config. & Eval. ceiling & Output \\
\midrule
\href{https://huggingface.co/Qwen/Qwen2.5-VL-3B-Instruct}{Qwen2.5-VL-3B-Instruct} & MRoPE & 8K $\to$ 32K\textsuperscript{a} & 128,000 & 128,000 & 1,024 \\
\href{https://huggingface.co/Qwen/Qwen2.5-VL-7B-Instruct}{Qwen2.5-VL-7B-Instruct} & MRoPE & 8K $\to$ 32K\textsuperscript{a} & 128,000 & 128,000 & 1,024 \\
\addlinespace[3pt]
\href{https://huggingface.co/Qwen/Qwen3-VL-2B-Instruct}{Qwen3-VL-2B-Instruct} & MRoPE-I & 8K $\to$ 256K\textsuperscript{b} & 262,144 & 262,144 & 1,024 \\
\href{https://huggingface.co/Qwen/Qwen3-VL-4B-Instruct}{Qwen3-VL-4B-Instruct} & MRoPE-I & 8K $\to$ 256K\textsuperscript{b} & 262,144 & 262,144 & 1,024 \\
\href{https://huggingface.co/Qwen/Qwen3-VL-8B-Instruct}{Qwen3-VL-8B-Instruct} & MRoPE-I & 8K $\to$ 256K\textsuperscript{b} & 262,144 & 262,144 & 1,024 \\
\href{https://huggingface.co/Qwen/Qwen3-VL-8B-Thinking}{Qwen3-VL-8B-Thinking} & MRoPE-I & 8K $\to$ 256K\textsuperscript{b} & 262,144 & 262,144 & 4,096 \\
\addlinespace[3pt]
\href{https://huggingface.co/OpenGVLab/InternVL3-2B-hf}{InternVL3-2B-HF} & Dynamic NTK & 32K\textsuperscript{c} & 32,768 & 65,536 & 4,096 \\
\href{https://huggingface.co/OpenGVLab/InternVL3-8B-hf}{InternVL3-8B-HF} & Dynamic NTK & 32K\textsuperscript{c} & 32,768 & 65,536 & 4,096 \\
\addlinespace[3pt]
\href{https://huggingface.co/modelscope/llava-llama3.1-8b}{LLaVA-Llama3.1-8B} & 1D (Llama3.1) & Not reported\textsuperscript{d} & 131,072 & 131,072 & 1,024 \\
\addlinespace[3pt]
\href{https://huggingface.co/llava-hf/llava-onevision-qwen2-7b-ov-hf}{LLaVA-OneVision-7B} & 1D (Qwen2) & 32K\textsuperscript{e} & 32,768 & 131,072 & 1,024 \\
\bottomrule\end{tabular}\end{table}\endgroup

\paragraph{Training-context sources.}
\textsuperscript{*}K denotes 1,024 tokens. Entries describe published family recipes or public training scripts, rather than checkpoint-specific training logs.
\textsuperscript{a}Qwen2.5-VL uses 8,192-token sequences in its first two pretraining stages and 32,768 in long-context pretraining~\citep{bai2025qwen25vl}.
\textsuperscript{b}Qwen3-VL progresses through 8,192, 32,768, and 262,144 tokens; its SFT recipe also progresses from 32K to 256K~\citep{bai2025qwen3vl}.
\textsuperscript{c}The InternVL3 report describes training infrastructure supporting sequences up to 32K; it does not separately specify the stage-wise training maxima of the evaluated 2B/8B releases~\citep{zhu2025internvl3}.
\textsuperscript{d}The LLaVA-Llama3.1 release describes multimodal alignment and instruction tuning but does not state a training sequence length~\citep{checkpoint_llava31_8b}.
\textsuperscript{e}The public OneVision \href{https://github.com/LLaVA-VL/LLaVA-NeXT/blob/main/scripts/train/finetune_si.sh}{single-image} and \href{https://github.com/LLaVA-VL/LLaVA-NeXT/blob/main/scripts/train/finetune_ov.sh}{one-vision} finetuning scripts set \texttt{model\_max\_length=32768}~\citep{li2024onevision}.

Training sequence length and visual evidence span are distinct: packing multiple examples into a long sequence does not imply that each image was trained with a question that far away. The 24K/32K text conditions thus probe visual access over controlled separations, while the reported training contexts locate those tests relative to each family's published recipe.

\subsection{Evaluation groups}
Table~\ref{tab:protocols} summarizes the question counts for task accuracy and consensus scores. The TextVQA length study uses the same questions as the main evaluation. The POPE and TextVQA length curves, placement study, and strength study use InternVL3-2B. The supplementary ChartQA length study and POPE layer-wise attention analysis use InternVL3-1B. Appendices~\ref{app:diagnostics} and~\ref{app:efficiency} describe attention measurements and timing protocols, respectively.

\begin{table}[htbp]\centering\small
\caption{Task evaluations. I/T denotes Instruct/Thinking. $K=1,024$.}
\label{tab:protocols}
\begin{tabular}{p{.19\linewidth}p{.57\linewidth}r}\toprule
Study & Conditions and models & Questions per score \\\midrule
Main tasks & Ten checkpoints, ten tasks, 32K & 500 \\
POPE lengths & Six checkpoints, 0/1/2/4/8/16/32K & 500 \\
TextVQA lengths & Six checkpoints, seven lengths & 500 \\
Five layouts & Six checkpoints, 32K total text, 0/8/16/24/32K after image & 500 \\
Strength & Six checkpoints, 32K, $\rho=0$ to 1.5 & 500 \\
Update schedule & Qwen2.5-7B, Qwen3-8B-I, InternVL3-2B, LLaVA. TextVQA/MathVista, 32K & 500 \\
Position-only controls & Four checkpoints, two shifts on short inputs & 500 \\
ChartQA lengths & Six checkpoints, seven lengths & 500 \\\bottomrule
\end{tabular}\end{table}

\subsection{Question selection and context construction}
The main results in Table~\ref{tab:main} use 500 questions per checkpoint and task, including CountBench. Each Vanilla/static comparison uses matching image questions and output budgets. All records contain a single image. POPE uses the adversarial split; TextVQA uses validation questions without OCR hints. Diagnostic and ablation results use the cohorts listed in Table~\ref{tab:protocols}.

The TextVQA length study evaluates the same 500 validation questions across all seven lengths and both methods.

The context corpus comprises 49 Paul Graham essays distributed with DIPE's evaluation resources. Files are sorted lexicographically, stripped of leading and trailing whitespace, flattened by replacing newlines with spaces, and concatenated. Each native tokenizer encodes this text; the first $L$ tokens are decoded into the prompt. Context text is drawn independently of the visual question and provides no designated task evidence. Retokenization and template boundaries can make the native position increment differ from $L$, so VPR derives its offset from prepared position IDs.

\subsection{Prompts and decoding}
The prompt contains the native system/user prefix, image, text context, question and options, and assistant-generation prefix. The short-context reference retains the image, question and template without the extended context. Qwen and LLaVA use native chat templates. InternVL uses the following evaluation system prompt:
\begin{quote}\small
You are an AI assistant that rigorously follows this response protocol:
1. First, conduct a detailed analysis of the question. Consider different angles, potential solutions, and reason through the problem step-by-step. Enclose this entire thinking process within \texttt{<think>} and \texttt{</think>} tags.
2. After the thinking section, provide a clear, concise, and direct answer to the user's question. Separate the answer from the think section with a newline.
Ensure that the thinking process is thorough but remains focused on the query. The final answer should be standalone and not reference the thinking section.
\end{quote}
This prompt does not designate InternVL as a separately released Thinking checkpoint. Qwen3 Thinking and Instruct are distinct releases. Task prompts preserve answer options and requests for the final answer, such as an option letter for AI2D or a concise response for ChartQA, DocVQA and GQA.

Decoding selects the argmax token and terminates at EOS or the output cap. A token loop exposes the cache intervention boundary. Experiments use independent single-GPU processes on H200 GPUs with BF16. The principal task and diagnostic environment uses PyTorch 2.8.0, CUDA 12.6 and Transformers 4.57.6; the schedule/cost environment is specified in Appendix~\ref{app:efficiency}. SDPA is used normally; explicit attention is used when needed for attention measurements and the weighting control.

\subsection{Task-specific metrics}
POPE extracts a yes/no answer using final-answer cues and supported negative forms. CountBench normalizes integers and English number words. ChartQA accepts numeric answers within 5\% relative error; a zero reference requires exact numeric agreement. Non-numeric answers use case-normalized string comparison. MathVista uses deterministic extraction appropriate to the record's answer type. AI2D and MMStar compare extracted options, while GQA and RealWorldQA use normalized short answers or options as appropriate.

For DocVQA, with reference set $\mathcal B$ and extracted prediction $\hat a$,
\begin{align}
d(\hat a,b)&=\frac{\operatorname{EditDistance}(\hat a,b)}{\max(|\hat a|,|b|)},\\
\operatorname{ANLS}(\hat a,\mathcal B)&=
\max_{b\in\mathcal B}\left[(1-d(\hat a,b))\mathbf1\{d(\hat a,b)<0.5\}\right].
\end{align}
The score is the sample mean multiplied by 100. ANLS permits partial transcription credit, so per-question score increases need not correspond to exact answers.

TextVQA uses normalized ten-answer EvalAI leave-one-out consensus without OCR hints. For each held-out annotator, the score is the minimum of one and the matching-answer count among the remaining nine divided by three. We average over annotators and questions, then multiply by 100. Extraction recognizes supported final-answer phrases, boxed answers and text following thinking blocks; no external language-model judge is used. Answer formatting and truncation can affect the measured task score, which motivates reporting generation behavior alongside extreme strength settings.

\subsection{Uniform attention amplification}
Attention$\times2$ adds $\log2$ to image logits before joint normalization:
\begin{equation}
s'_i=s_i+\mathbf1[i\in\imgset]\log2.
\label{eq:attwo}
\end{equation}
At fixed queries and keys, the common multiplier cancels when attention is normalized within the image, preserving the relative weights of its visual tokens. Rephasing instead changes token-specific logits.

\section{Supplementary Diagnostics and Ablations}
\label{app:diagnostics}
\subsection{POPE recovery across context lengths}
\label{app:updated_scans}
\begin{figure}[htbp]\centering
\includegraphics[width=.96\linewidth]{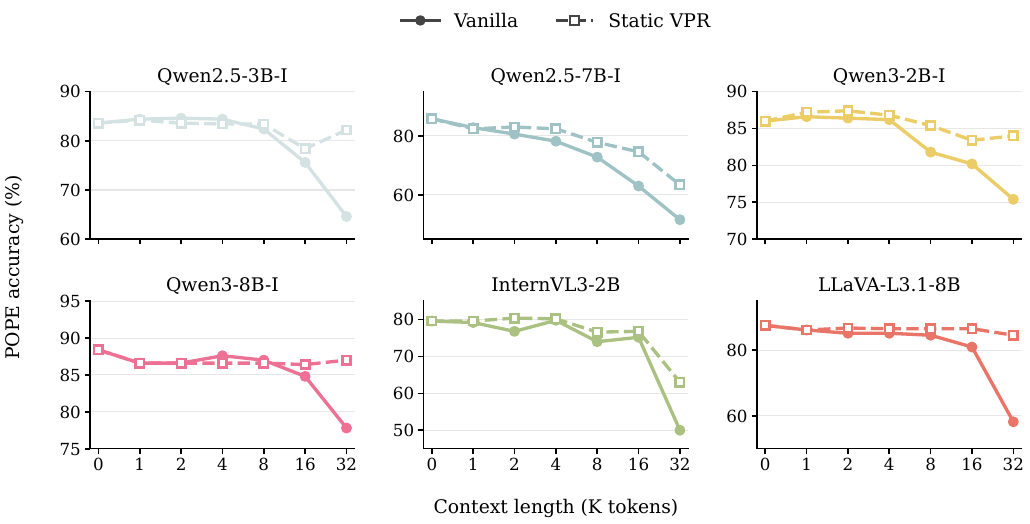}
\caption{POPE accuracy across seven context lengths. Solid lines show Vanilla; dashed lines show static VPR. Each panel uses its own vertical scale.}
\label{fig:length_recovery}\end{figure}

Figure~\ref{fig:length} shows how Vanilla accuracy changes as more text separates the image from the question. Figure~\ref{fig:length_recovery} adds static VPR to the same POPE length scan. At 0K there is no added separation to compensate for, so the two curves share a starting point.

\subsection{ChartQA across context lengths}
\label{app:chartqa_lengths}
\begin{figure}[htbp]\centering
\includegraphics[width=.96\linewidth]{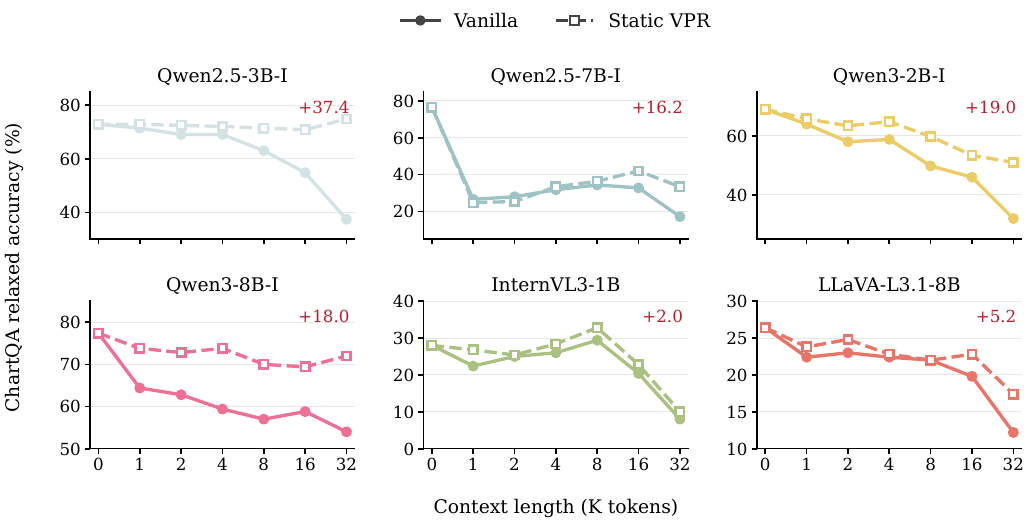}
\caption{ChartQA accuracy across seven context lengths. Panels use model-specific vertical ranges; red labels show 32K gains in percentage points.}
\label{fig:chartqa_lengths}\end{figure}

Figure~\ref{fig:chartqa_lengths} shows ChartQA accuracy at seven context lengths. Across the 36 nonzero-length comparisons, VPR improves 33, ties in one and decreases in two. At 32K, all six models improve, with gains ranging from 2.0 to 37.4 percentage points. Qwen3-8B-Instruct improves from 54.0\% to 72.0\%, while Qwen2.5-3B improves from 37.4\% to 74.8\%. The two decreases occur for Qwen2.5-7B at 1K and 2K. These trajectories complement TextVQA: re-anchoring improves chart understanding across models at long separation, while its effect at shorter lengths varies. Qwen2.5-7B drops from 76.6\% at 0K to 26.6\% at 1K; the available records do not identify a single cause.

\subsection{Position-only control protocol}
\label{app:position_control}
Figure~\ref{fig:position_control} uses the same short image--question input across conditions for each example. All four checkpoints share 500 questions per task, drawn from 84 POPE images and 478 TextVQA images. Both tasks cover nominal shifts of 8, 16, 24 and 32K. For each nominal length $L$, native prompt preparation supplies the extra image--question displacement $\Delta_L$ of the corresponding long-input setting. Preparing this reference determines positions; its text is not passed through the model in either control.

The question-shift arm adds $\Delta_L$ to the native positions of the question-processing segment, including the assistant prefix; generated positions continue from that shifted suffix. The visual-key arm instead rephases cached image keys by $-\Delta_L$ before question prefill, leaving text positions and all values unchanged. Both preserve physical token counts and causal visibility. The latter can assign negative effective positions to visual keys; these are rotary phases, not negative cache indices. The arms affect different interactions: shifting the question segment changes its relation to the entire preceding prefix, whereas rephasing visual keys changes access to those keys specifically. They are therefore complementary controls, not equivalent implementations of an isolated scalar distance change.

Each arm is compared with its own matching short-input run, not a reference imported from another evaluation cohort. Each checkpoint retains its native preprocessing and uses greedy decoding with the same output cap in both arms (4,096 tokens for InternVL3-2B and 1,024 for the others). We retain capped and unparsed outputs in the denominator. These measurements establish sensitivity to positional intervention without extra text keys; they do not identify whether the response arises from particular rotary frequencies or from positional relations rarely encountered in training.

\subsection{POPE attention across lengths and layers}
\label{app:pope_attention}
\begin{figure}[htbp]\centering
\includegraphics[width=.96\linewidth]{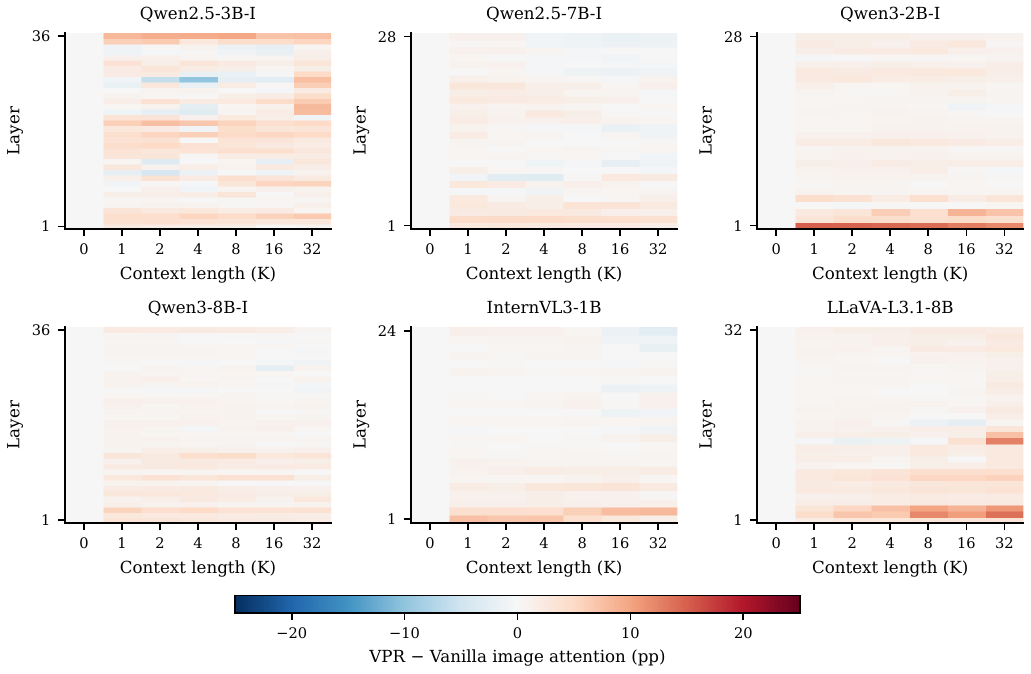}
\caption{Layer-wise change in image-attention mass, in percentage points, using a shared color scale. Layer indices are model-local and decoder depths differ.}
\label{fig:attention_layers}\end{figure}

Attention uses 100 common questions from 17 images at the prompt query predicting the first answer token. We sum attention over visual keys and average over heads, layers and questions. Figure~\ref{fig:attention_layers} shows the layer-wise change induced by VPR. Image-cluster intervals for the aggregate means use 2,000 bootstrap replicates. Questions sharing an image are not treated as independent visual scenes.

\subsection{TextVQA attention}
\label{app:textvqa_attention}
The TextVQA mechanism study uses 100 questions from 100 distinct images, shared by Qwen2.5-7B, Qwen3-2B-Instruct and LLaVA-Llama3.1-8B. We evaluate Vanilla and static VPR at 0, 8K and 32K; at zero length the intervention is the identity. Attention is measured at the final prompt query, before the first answer token. We normalize over all available keys, sum over image tokens and average over query heads. Layer summaries average over the same questions; aggregate mass additionally averages over layers. Thus the measurement retains competition from text keys rather than renormalizing within the image.

\begin{figure}[htbp]\centering
\includegraphics[width=.96\linewidth]{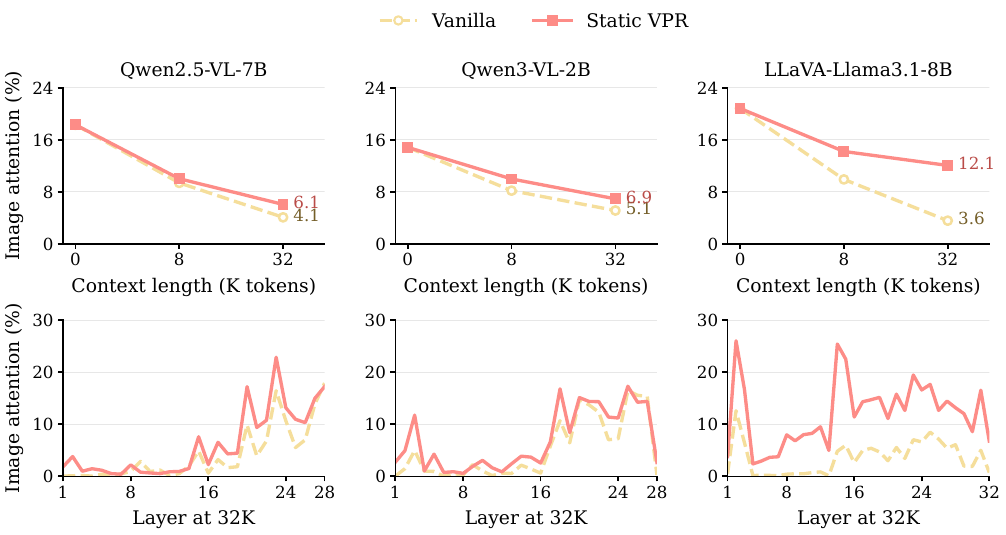}
\caption{TextVQA image-attention mass across lengths (top) and layers at 32K (bottom). The same image questions are used throughout.}
\label{fig:textvqa_attention_profiles}\end{figure}

Figure~\ref{fig:textvqa_attention_profiles} summarizes the length and layer profiles. At 32K, mean image-attention mass rises from 4.07\% to 6.05\% for Qwen2.5-7B, from 5.12\% to 6.91\% for Qwen3-2B, and from 3.55\% to 12.05\% for LLaVA. Increases occur in 22 of 28, 25 of 28, and all 32 layers, respectively. Question-wise Pearson correlations between attention and score changes are 0.366, 0.236 and 0.168. Attention allocation and answer quality are related observables, not interchangeable metrics. These masses use a different cohort from the POPE diagnostic and measure total visual attention, whereas Figure~\ref{fig:evidence_regions} measures its distribution within the image.

\subsection{Attention within annotated evidence regions}
\label{app:evidence_regions}
All six checkpoints in Figure~\ref{fig:evidence_regions} use the same 100 distinct TextVQA validation images. We select one question per image, require agreement among at least six of its ten reference answers, and match the consensus answer to a unique TextOCR region with valid coordinates. Selection uses only the annotations. For each image, Vanilla and static \vpr\ ($\rho=1$) receive identical inputs with 32,768 intervening text tokens. We read attention at the final prompt query, immediately before answer generation.

For layer $\ell$ and query head $h$, let $a_i^{\ell,h}$ denote attention to key $i$ after normalization over all visible keys. For each visual token, $w_i\in[0,1]$ is the fraction of its spatial footprint covered by the union of the answer boxes. We map boxes through each checkpoint's native preprocessing, accounting for grid merging in Qwen, tiling and pixel shuffling in InternVL, and feature packing and unpadding in LLaVA. Padding contributes no evidence. Non-spatial visual keys have $w_i=0$ and contribute to total visual attention. For the visual key set $\imgset$, define
\begin{align}
M_{\ell,h} &= \sum_{i\in\imgset} a_i^{\ell,h}, &
R_{\ell,h} &= \sum_{i\in\imgset} w_i a_i^{\ell,h}, \notag\\
G_{\ell,h} &= R_{\ell,h}/M_{\ell,h}.
\label{eq:evidence_share}
\end{align}
For each image $x$, we average the ratio over the $H_\ell$ query heads in each layer and then over the $L$ layers:
\begin{equation}
G_x=\frac{1}{L}\sum_{\ell=1}^{L}\frac{1}{H_\ell}\sum_{h=1}^{H_\ell}G_{\ell,h}.
\end{equation}
We apply log-softmax in FP64 to the final-query FP32 logits and evaluate the ratio in log space, retaining all heads even when their visual attention mass is very small.

The bars report $100\,\overline{G_x}$ for each condition. To estimate the gain, we first subtract the Vanilla share from the static \vpr\ share for each image. We resample image indices with replacement 10,000 times (seed 20261103), using the same draws for all conditions and checkpoints. The 2.5th and 97.5th percentiles give the intervals for each condition mean and, separately, for each paired mean difference. The boxes identify text matching the reference answer and can omit surrounding context useful for interpreting it.

\subsection{Strength and answer production}
The strength study uses output caps of 1,024 tokens for Qwen and LLaVA and 4,096 for InternVL3-2B. All outputs remain in the score denominator, including capped and unparsed answers.

\paragraph{Beyond-reference positions.}
For $\Delta>0$, Equation~\ref{eq:distance} gives $\widetilde d_i<0$ when $\rho>1+d_i^{(0)}/\Delta$. Thus $\rho>1$ does not by itself imply a sign reversal for every image token. With $\Delta\approx32$K, strengths 1.25 and 1.5 subtract approximately 8K and 16K from the reference separation. The exact offsets follow native position IDs; multidimensional encodings require coordinate-wise interpretation. A negative signed separation changes the effective positional relation without revealing future tokens. A position within a checkpoint's configured range is not evidence that this relation was encountered during multimodal training. Table~\ref{tab:models} distinguishes configuration and training context; InternVL3 additionally uses dynamic NTK scaling. We therefore treat the rightmost columns as exploratory settings, not as a controlled isolation of positional extrapolation.

\subsection{Joint image and text lookup}
\label{app:joint_mapping}
For each of 200 TextVQA images, two text mappings assign a four-letter answer code to the relevant visual detail, yielding 400 answers per model, length and method. We evaluate 8K, 16K, 24K and 32K text contexts. Table~\ref{tab:joint_mapping} reports code accuracy averaged over both mappings. A response is correct only if its extracted code matches the target; unparsed and capped responses remain in the denominator. The static schedule uses $\rho=1$ before question processing. Qwen2.5's blank-image control at 32K scores 12.5\%, compared with 11.0\% for full-input Vanilla. This control provides no positive separation for Vanilla and limits conclusions about this checkpoint's baseline use of visual evidence.

\section{Update Schedules and Computational Cost}
\subsection{Update-schedule protocol}
\label{app:schedule}
The MathVista continuous scores come from the 500-question update-schedule evaluation. The continuous TextVQA evaluation uses 500 questions from 478 images, shared by all four checkpoints. We use greedy decoding with normal EOS stopping and output caps of 1,024 tokens for Qwen and LLaVA and 4,096 for InternVL.

Static compensation retains $\delta=\Delta$, while continuous compensation uses $\delta_j=\Delta+j$, with $j=0$ predicting the first answer token. Both reconstruct target phases from saved original visual keys. InternVL retains native dynamic NTK factor two and request-level frequency resets.

\subsection{Complete efficiency measurements}
\label{app:efficiency}
Table~\ref{tab:effall} covers four models, 8K/32K input text, 128/512 fixed output tokens and three methods. The timing runtime uses H200 GPUs, BF16, SDPA, PyTorch 2.5.0a0 (NVIDIA 24.10), CUDA 12.6 and Transformers 4.46.2 with the supplied adapters. The eight question records reference four distinct images. Each sample and condition receives one warmup and five measured repetitions, with matched conditions interleaved on the same H200 GPU. This produces 1,920 formal measurements plus 160 native-path controls. Output generation ignores EOS to hold work fixed; these answers do not enter task scoring. Timing uncertainty resamples the four image clusters, retaining all associated questions, five nested repetitions and paired conditions, using 2,000 replicates and seed 20261101.

Inference time starts before prefix prefill and ends after the final token is selected and CUDA work is synchronized. It includes frequency reset, saved-key copies, phase construction, cache writes, question prefill and decoding. Model loading, input preprocessing, output string decoding and file writes are excluded. Shared input preparation is measured separately and added once to obtain total time. TTFT uses the same inference start and ends at the first selected token; decode throughput is $(B-1)/t_{\mathrm{decode}}$. Overhead is the ratio of condition means minus one, relative to the matched split-prefill Vanilla. The complete CSV release includes preprocessing, total time and all interval endpoints.

Peak allocated memory is reset per request and includes weights, input tensors, cache and temporary allocations. Table~\ref{tab:effall} uses GiB ($2^{30}$ bytes). Static and continuous peak allocated memory equal the split baseline in the measured cells; this does not imply the absence of additional allocations below that peak. Small negative timing differences are retained as measurements.
\begingroup\small\setlength{\tabcolsep}{4pt}
\begin{longtable}{lrrlrrlrrr}
\caption{Complete inference measurements: time and TTFT in seconds, overhead relative to split Vanilla in percent, decode rate in tokens/s, peak allocated memory in GiB.}\label{tab:effall} \\
\toprule Model & Input K & Output & Method & Time & Overhead & 95\% interval & TTFT & Rate & Peak \\\midrule\endfirsthead
\toprule Model & Input K & Output & Method & Time & Overhead & 95\% interval & TTFT & Rate & Peak \\\midrule\endhead
\bottomrule\endfoot
InternVL3-2B & 8 & 128 & Vanilla & 1.818 & +0.00 & [+0.00, +0.00] & 0.107 & 74.3 & 4.716 \\
InternVL3-2B & 8 & 128 & Continuous & 2.459 & +35.29 & [+34.82, +35.66] & 0.115 & 54.2 & 4.716 \\
InternVL3-2B & 8 & 128 & Static & 1.817 & -0.06 & [-0.38, +0.27] & 0.115 & 74.7 & 4.716 \\
InternVL3-2B & 8 & 512 & Vanilla & 7.001 & +0.00 & [+0.00, +0.00] & 0.107 & 74.2 & 4.716 \\
InternVL3-2B & 8 & 512 & Continuous & 9.571 & +36.71 & [+36.25, +37.15] & 0.115 & 54.1 & 4.716 \\
InternVL3-2B & 8 & 512 & Static & 7.004 & +0.04 & [-0.16, +0.22] & 0.115 & 74.2 & 4.716 \\
InternVL3-2B & 32 & 128 & Vanilla & 2.283 & +0.00 & [+0.00, +0.00] & 0.549 & 73.3 & 6.896 \\
InternVL3-2B & 32 & 128 & Continuous & 2.941 & +28.82 & [+28.18, +30.43] & 0.561 & 53.5 & 6.896 \\
InternVL3-2B & 32 & 128 & Static & 2.934 & +28.51 & [+27.67, +28.84] & 0.562 & 53.6 & 6.896 \\
InternVL3-2B & 32 & 512 & Vanilla & 7.484 & +0.00 & [+0.00, +0.00] & 0.548 & 73.7 & 6.896 \\
InternVL3-2B & 32 & 512 & Continuous & 10.019 & +33.87 & [+33.51, +34.72] & 0.561 & 54.1 & 6.896 \\
InternVL3-2B & 32 & 512 & Static & 10.015 & +33.82 & [+33.37, +34.81] & 0.561 & 54.1 & 6.896 \\
LLaVA-L3.1-8B & 8 & 128 & Vanilla & 2.348 & +0.00 & [+0.00, +0.00] & 0.403 & 65.3 & 18.047 \\
LLaVA-L3.1-8B & 8 & 128 & Continuous & 2.623 & +11.71 & [+11.20, +12.74] & 0.406 & 57.4 & 18.047 \\
LLaVA-L3.1-8B & 8 & 128 & Static & 2.349 & +0.03 & [-0.27, +0.31] & 0.406 & 65.4 & 18.047 \\
LLaVA-L3.1-8B & 8 & 512 & Vanilla & 8.269 & +0.00 & [+0.00, +0.00] & 0.403 & 65.0 & 18.047 \\
LLaVA-L3.1-8B & 8 & 512 & Continuous & 9.305 & +12.53 & [+12.05, +13.00] & 0.406 & 57.5 & 18.047 \\
LLaVA-L3.1-8B & 8 & 512 & Static & 8.263 & -0.07 & [-0.31, +0.21] & 0.406 & 65.1 & 18.047 \\
LLaVA-L3.1-8B & 32 & 128 & Vanilla & 3.991 & +0.00 & [+0.00, +0.00] & 1.993 & 63.6 & 23.779 \\
LLaVA-L3.1-8B & 32 & 128 & Continuous & 4.251 & +6.52 & [+6.37, +6.67] & 1.996 & 56.3 & 23.779 \\
LLaVA-L3.1-8B & 32 & 128 & Static & 3.997 & +0.14 & [+0.08, +0.18] & 1.996 & 63.5 & 23.779 \\
LLaVA-L3.1-8B & 32 & 512 & Vanilla & 10.011 & +0.00 & [+0.00, +0.00] & 1.993 & 63.7 & 23.779 \\
LLaVA-L3.1-8B & 32 & 512 & Continuous & 11.051 & +10.39 & [+10.24, +10.66] & 1.993 & 56.4 & 23.779 \\
LLaVA-L3.1-8B & 32 & 512 & Static & 10.023 & +0.12 & [+0.04, +0.35] & 1.994 & 63.7 & 23.779 \\
Qwen2.5-7B & 8 & 128 & Vanilla & 2.259 & +0.00 & [+0.00, +0.00] & 0.306 & 65.2 & 17.085 \\
Qwen2.5-7B & 8 & 128 & Continuous & 2.539 & +12.39 & [+10.46, +13.04] & 0.308 & 57.2 & 17.085 \\
Qwen2.5-7B & 8 & 128 & Static & 2.278 & +0.84 & [-0.01, +1.60] & 0.307 & 64.7 & 17.085 \\
Qwen2.5-7B & 8 & 512 & Vanilla & 8.043 & +0.00 & [+0.00, +0.00] & 0.306 & 66.1 & 17.085 \\
Qwen2.5-7B & 8 & 512 & Continuous & 9.013 & +12.06 & [+11.12, +12.38] & 0.307 & 58.7 & 17.085 \\
Qwen2.5-7B & 8 & 512 & Static & 8.035 & -0.09 & [-0.96, +0.32] & 0.307 & 66.1 & 17.085 \\
Qwen2.5-7B & 32 & 128 & Vanilla & 3.465 & +0.00 & [+0.00, +0.00] & 1.537 & 65.9 & 21.692 \\
Qwen2.5-7B & 32 & 128 & Continuous & 3.709 & +7.03 & [+6.71, +7.21] & 1.538 & 58.5 & 21.692 \\
Qwen2.5-7B & 32 & 128 & Static & 3.470 & +0.13 & [+0.01, +0.21] & 1.538 & 65.8 & 21.692 \\
Qwen2.5-7B & 32 & 512 & Vanilla & 9.416 & +0.00 & [+0.00, +0.00] & 1.537 & 64.9 & 21.692 \\
Qwen2.5-7B & 32 & 512 & Continuous & 10.443 & +10.91 & [+10.75, +11.07] & 1.538 & 57.4 & 21.692 \\
Qwen2.5-7B & 32 & 512 & Static & 9.447 & +0.33 & [-0.02, +0.69] & 1.536 & 64.6 & 21.692 \\
Qwen3-8B-I & 8 & 128 & Vanilla & 3.131 & +0.00 & [+0.00, +0.00] & 0.361 & 45.9 & 18.386 \\
Qwen3-8B-I & 8 & 128 & Continuous & 3.458 & +10.44 & [+8.83, +12.35] & 0.363 & 41.2 & 18.386 \\
Qwen3-8B-I & 8 & 128 & Static & 3.156 & +0.80 & [-0.13, +1.92] & 0.363 & 45.5 & 18.386 \\
Qwen3-8B-I & 8 & 512 & Vanilla & 11.428 & +0.00 & [+0.00, +0.00] & 0.361 & 46.2 & 18.386 \\
Qwen3-8B-I & 8 & 512 & Continuous & 12.588 & +10.15 & [+9.30, +10.62] & 0.362 & 41.8 & 18.386 \\
Qwen3-8B-I & 8 & 512 & Static & 11.425 & -0.02 & [-0.85, +0.47] & 0.362 & 46.2 & 18.386 \\
Qwen3-8B-I & 32 & 128 & Vanilla & 4.738 & +0.00 & [+0.00, +0.00] & 2.018 & 46.7 & 24.211 \\
Qwen3-8B-I & 32 & 128 & Continuous & 5.033 & +6.23 & [+6.11, +6.40] & 2.019 & 42.2 & 24.211 \\
Qwen3-8B-I & 32 & 128 & Static & 4.741 & +0.06 & [+0.01, +0.17] & 2.018 & 46.7 & 24.211 \\
Qwen3-8B-I & 32 & 512 & Vanilla & 12.940 & +0.00 & [+0.00, +0.00] & 2.016 & 46.8 & 24.211 \\
Qwen3-8B-I & 32 & 512 & Continuous & 14.121 & +9.12 & [+9.04, +9.20] & 2.018 & 42.2 & 24.211 \\
Qwen3-8B-I & 32 & 512 & Static & 12.969 & +0.22 & [+0.09, +0.39] & 2.018 & 46.7 & 24.211 \\
\end{longtable}\endgroup

\paragraph{Native execution path.}
The native control prefills the identical full tokenized prompt in one standard model forward, then uses the same manual greedy KV decoding loop as the split conditions. It does not use the \texttt{generate} scheduler. Table~\ref{tab:native_cost} compares these paths at 32K/128 tokens. The split/native difference includes actual execution and native frequency evolution; identical input tokens do not require bitwise-identical output sequences. Static versus native costs 2.29\% on both Qwen models, 3.50\% on LLaVA and 31.15\% on InternVL.
\begingroup\small\renewcommand{\arraystretch}{1.1}\setlength{\tabcolsep}{4pt}
\begin{longtable}{llrrrr}
\caption{Complete-prefill native-path control at 32K and 128 output tokens. Time in seconds; overhead and intervals in percent.}\label{tab:native_cost} \\
\toprule Model & Contrast & Method time & Native time & Overhead & 95\% interval \\\midrule\endfirsthead
\toprule Model & Contrast & Method time & Native time & Overhead & 95\% interval \\\midrule\endhead
\bottomrule\endfoot
Qwen2.5-7B & split\_vs\_native & 3.465 & 3.392 & +2.16 & [+2.07, +2.38] \\
Qwen2.5-7B & static\_vs\_native & 3.470 & 3.392 & +2.29 & [+2.18, +2.45] \\
Qwen2.5-7B & continuous\_vs\_native & 3.709 & 3.392 & +9.34 & [+9.27, +9.51] \\
Qwen3-8B-I & split\_vs\_native & 4.738 & 4.635 & +2.22 & [+2.15, +2.45] \\
Qwen3-8B-I & static\_vs\_native & 4.741 & 4.635 & +2.29 & [+2.18, +2.56] \\
Qwen3-8B-I & continuous\_vs\_native & 5.033 & 4.635 & +8.59 & [+8.44, +8.88] \\
InternVL3-2B & split\_vs\_native & 2.283 & 2.237 & +2.06 & [+1.54, +2.88] \\
InternVL3-2B & static\_vs\_native & 2.934 & 2.237 & +31.15 & [+30.75, +31.56] \\
InternVL3-2B & continuous\_vs\_native & 2.941 & 2.237 & +31.47 & [+30.42, +33.91] \\
LLaVA-L3.1-8B & split\_vs\_native & 3.991 & 3.862 & +3.36 & [+2.43, +4.10] \\
LLaVA-L3.1-8B & static\_vs\_native & 3.997 & 3.862 & +3.50 & [+2.61, +4.21] \\
LLaVA-L3.1-8B & continuous\_vs\_native & 4.251 & 3.862 & +10.09 & [+9.24, +10.75] \\
\end{longtable}\endgroup

\paragraph{Frequency and cache updates.}
Table~\ref{tab:schedule_updates} distinguishes native frequency changes from logical updates of all visual-key cache entries. Qwen and LLaVA keep a fixed frequency basis: static uses one cache update, while continuous uses one per generated token. InternVL at 32K changes its basis during question prefill and generation, requiring 129/513 static updates for 128/512 generated tokens. At 8K no such frequency changes occur and static uses one update. This explains why fixed displacement does not guarantee a single cache write under dynamic RoPE.
\begin{table}[htbp]\centering\small
\renewcommand{\arraystretch}{1.1}
\caption{Native frequency changes and visual-cache updates per request. Counts are identical for Qwen2.5-7B, Qwen3-8B-Instruct and LLaVA-Llama3.1-8B. A logical cache update writes visual keys across all layers; Vanilla performs no VPR cache updates.}
\label{tab:schedule_updates}
\setlength{\tabcolsep}{7pt}
\begin{tabular}{@{}lrrr rr@{}}\toprule
& & & Native frequency & \multicolumn{2}{c}{Cache updates} \\
\cmidrule(lr){5-6}
Models & Input K & Output & changes & Static & Continuous \\\midrule
Qwen2.5 / Qwen3 / LLaVA & 8 / 32 & 128 & 0 & 1 & 128 \\
& 8 / 32 & 512 & 0 & 1 & 512 \\\addlinespace[3pt]
InternVL3-2B & 8 & 128 & 0 & 1 & 128 \\
& 8 & 512 & 0 & 1 & 512 \\
& 32 & 128 & 129 & 129 & 129 \\
& 32 & 512 & 513 & 513 & 513 \\\bottomrule
\end{tabular}
\end{table}

\paragraph{Reproducibility and release.}
Evaluation records specify checkpoint identities, question IDs, templates, image preprocessing, context construction, output budgets and answer extraction. The supporting materials include configuration records, implementation and scoring excerpts, summary results and representative responses.

\end{document}